\documentclass[sigconf]{acmart}
\AtBeginDocument{%
  }

\setcopyright{acmlicensed}

\copyrightyear{2025}
\acmYear{2025}
\setcopyright{acmlicensed}\acmConference[KDD '25]{Proceedings of the 31st ACM SIGKDD Conference on Knowledge Discovery and Data Mining V.2}{August 3--7, 2025}{Toronto, ON, Canada}
\acmBooktitle{Proceedings of the 31st ACM SIGKDD Conference on Knowledge Discovery and Data Mining V.2 (KDD '25), August 3--7, 2025, Toronto, ON, Canada}
\acmDOI{10.1145/3711896.3737139}
\acmISBN{979-8-4007-1454-2/2025/08}

\usepackage{url}
\usepackage{booktabs}
\usepackage{multirow}    
\usepackage{pifont}      
\usepackage{tablefootnote} 
\usepackage{enumitem}
\usepackage{subfigure}
\usepackage{algorithm}
\usepackage{algorithmic}
\usepackage{wrapfig}

\usepackage{colortbl}  
\definecolor{gray94}{gray}{.94}
\definecolor{gray90}{gray}{.90}
\newcommand{\grow}[1]{\rowcolor{gray94}{#1}}

\begin{document}

\title{Surface-based Molecular Design with Multi-modal Flow Matching}

\author{Fang Wu}
\email{fangwu97@stanford.edu}
\affiliation{%
  \institution{Stanford University}
  \city{Stanford}
  \state{CA}
  \country{USA}
}

\author{Zhengyuan Zhou}
\email{zzhou@ucsd.edu}
\affiliation{%
  \institution{University of California, San Diego}
  \city{La Jolla}
  \state{CA}
  \country{USA}
}

\author{Shuting Jin}
\email{shutingjin@wust.edu.cn}
\affiliation{%
  \institution{Wuhan University of Science and Technology}
  \city{Wuhan}
  \state{Hubei}
  \country{China}
}

\author{Xiangxiang Zeng}
\email{xzeng@foxmail.com}
\affiliation{%
  \institution{Hunan University}
  \city{Changsha}
  \state{Hunan}
  \country{China}
}

\author{Jure Leskovec}
\email{jure@cs.stanford.edu}
\affiliation{%
  \institution{Stanford University}
  \city{Stanford}
  \state{CA}
  \country{USA}
}

\author{Jinbo Xu$^\dagger$}
\email{jinboxu@moleculemind.com}
\affiliation{%
  \institution{MoleculeMind}
  \city{Beijing}
  \country{China}
}

\renewcommand{\shortauthors}{Fang Wu et al.}

\begin{abstract}
    Therapeutic peptides show promise in targeting previously undruggable binding sites, with recent advancements in deep generative models enabling full-atom peptide co-design for specific protein receptors. However, the critical role of molecular surfaces in protein-protein interactions (PPIs) has been underexplored. To bridge this gap, we propose an \emph{omni-design} peptides generation paradigm, called SurfFlow, a novel surface-based generative algorithm that enables comprehensive co-design of sequence, structure, and surface for peptides. 
    SurfFlow employs a multi-modality conditional flow matching (CFM) architecture to learn distributions of surface geometries and biochemical properties, enhancing peptide binding accuracy. Evaluated on the comprehensive PepMerge benchmark, SurfFlow consistently outperforms full-atom baselines across all metrics. These results highlight the advantages of considering molecular surfaces in \emph{de novo} peptide discovery and demonstrate the potential of integrating multiple protein modalities for more effective therapeutic peptide discovery.
\end{abstract}


\maketitle

\section{Introduction}
Peptides, short-chain proteins composed of roughly 2 to 50 amino acids, play critical roles in various biological processes, including cell signaling, enzymatic catalysis, and immune responses~\citep{wang2022therapeutic}.
They are essential mediators in pharmacology due to their ability to bind cell surface receptors with high affinity and specificity, such as intracellular effects with low toxicity, minimal immunogenicity, and ease of delivery~\citep{muttenthaler2021trends}. Conventional simulation or searching methods of peptide discovery rely on frequent calculations of physical energy functions, a process hindered by the vast design space~\citep{bhardwaj2016accurate}. This spurs a growing demand for computational approaches to facilitate \emph{in silico} peptide design and analysis.

Recently, diffusion~\citep{ho2020denoising} and flow~\citep{lipman2022flow} models illuminate tremendous promise in molecular design~\citep{guan20233d}, antibody engineering~\citep{wu2024hierarchical,martinkus2024abdiffuser}, protein design~\citep{yim2023fast,wu2024protein} as well as peptide discovery~\citep{ramasubramanian2024hybrid,wu2024d}. Unbound peptides often exist in high-energy, high-entropy states with unstable conformations, only becoming functional upon binding to target receptors. Thus, peptide design must be explicitly conditioned on binding pockets~\cite{vanhee2011computational}. Besides, as residues interact with each other through non-covalent forces formed by side-chain groups, increasing efforts are being made to capture protein-peptide interactions through full-atom geometries~\citep{kong2024full,lin2024ppflow}, pushing sequence-structure co-design beyond just backbones.
\begin{figure}[t]
    \centering
    \includegraphics[width=1\linewidth]{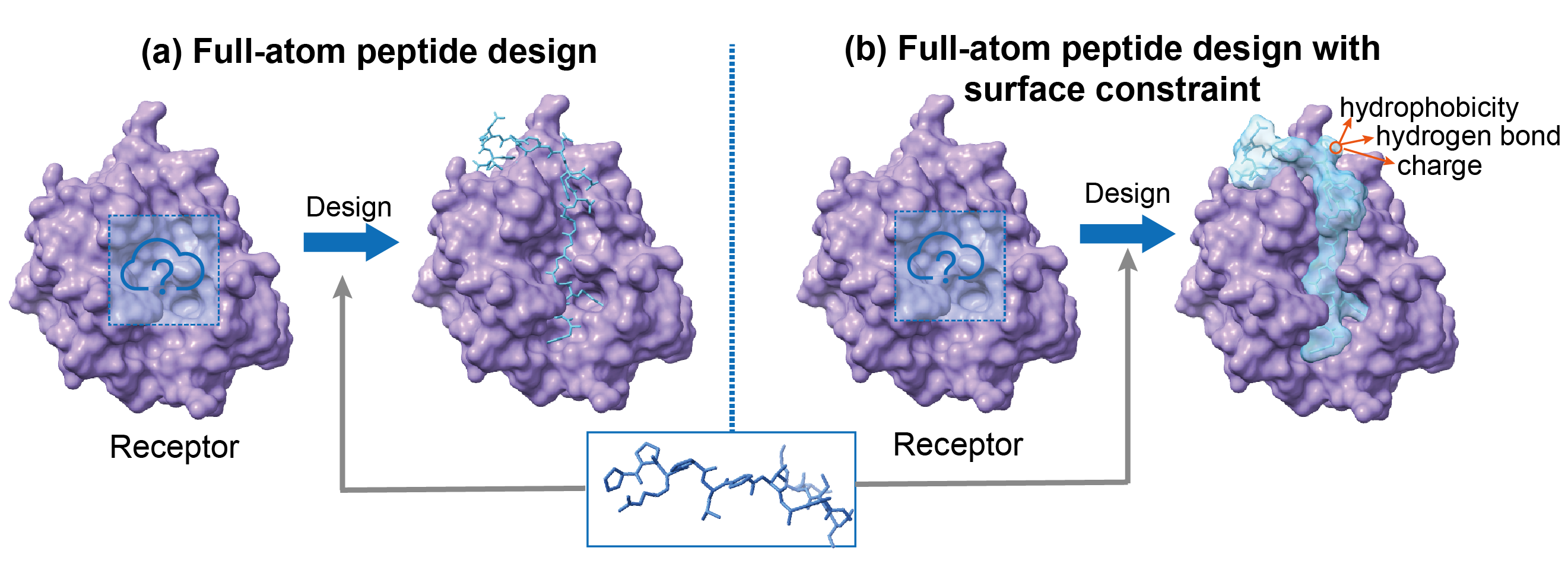}
    \caption{Comparison of full-atom peptide design with and without the surface constraint.}
    \Description{ }
    \label{fig:protein_design}
    \vspace{-1em}
\end{figure}

Meanwhile, growing attention is being given to molecular surfaces in the study of protein-protein interactions (PPIs), as PPIs are largely dictated by how the surfaces of interacting proteins complement each other~\citep {kastritis2013binding,song2024surfpro}. The surface's electrostatic potential and hydrophobicity are key determinants of PPIs' strength and specificity~\citep{jones1996principles,lee2023pre,lai2024interformer}, and its geometries, such as protrusions, grooves, and clefts, enable lock-and-key~\citep{morrison2006lock} or induced-fit mechanisms essential for specific binding. Those surfaces act as a fundamental interface that dictates how proteins recognize and bind to each other. For these reasons, it is vital to simultaneously consider all molecular modalities — sequence, structure, and surface — during peptide generation (see Fig.~\ref{fig:protein_design}), thereby enhancing consistency across aspects in what we term \emph{omni-design}. 

Toward this goal, we propose a novel omni-design generative algorithm dubbed SurfFlow. It applies multi-modality flow matching (FM)~\citep{lipman2022flow,albergo2022building,albergo2023stochastic} to internal structures and molecular surfaces, which are represented by surface point positions and unit norm vectors as a rigid frame in $\mathrm{SE}(3)$. Since complementary surface geometries alone do not guarantee successful binding-accurate and placement of charges, polarity, and hydrophobicity at the binding interface is also necessary~\citep{gainza2023novo}, SurfFlow incorporates those biochemical property constraints. Specifically, it exploits a discrete FM (DFM)~\citep{campbell2024generative} to perform on the discrete data space of some categorical surface features using Continuous-Time Markov Chains (CTMC). 
Moreover, considering the challenges of capturing irregular surface geometries, multiscale features, and inter-protein interactions in a scalable manner, we propose an equivariant surface geometric network (ESGN), which dynamically models heterogeneous surface graphs while uniquely incorporating both intra- and inter-surface interactions.
Finally, recognizing that key peptide attributes like cyclicity and disulfide bonds influence stability and binding affinity~\citep{buckton2021cyclic}, we include these factors as additional conditions to enhance the capacity and generalization of SurfFlow. We evaluate SurfFlow on the comprehensive peptide design benchmark PepMerge~\citep{li2024full}, and experiments demonstrate that it consistently outperforms full-atom baselines across all metrics, highlighting the advantages of considering surfaces for \emph{de novo} peptide design. 

\section{Preliminary and Background}
\paragraph{Proteins and Molecular Surfaces.} 
A protein consists of multiple amino acids, each defined by its type, backbone frame, and side-chain torsion angles~\citep{fisher2001lehninger}. The $i$-th residue type, denoted by $a_i \in \{1 \ldots 20\}$, is determined by its side-chain R group. The rigid frame of each residue is constructed from the coordinates of 4 backbone heavy atoms $\mathrm{N}$, $\mathrm{C}\alpha$, $\mathrm{C}$, and $\mathrm{O}$, with $\mathrm{C}\alpha$ positioned at the origin. This frame is represented by a position vector $\boldsymbol{x}_i \in \mathbb{R}^3$ and a rotation matrix $O_i \in \mathrm{SO}(3)$~\citep{jumper2021highly}. Unlike backbones, side-chains are more flexible, involving up to 4 rotatable torsion angles between side-chain atoms, denoted by $\boldsymbol{\chi}_i \in[0,2 \pi)^4$. Additionally, the backbone torsion angle $\varphi_i \in[0,2 \pi)$ affects the position of the oxygen atom.

We further consider the molecular surface, computed by moving a probe of a certain radius (approximately 1 Å) along the protein to calculate the Solvent Accessible Surface (SAS) and Solvent Excluded Surface (SES). The probe's coordinates define the surface as an oriented point cloud $Q = \{q_i: 1 \leq i \leq m\}$. Each point $q_i$ has associated attributes $(\boldsymbol{x}_i^s, \boldsymbol{n}_i^s, \boldsymbol{\tau}_i^s, \boldsymbol{\Upsilon}_i^s)$, where $\boldsymbol{x}_i^s \in \mathbb{R}^3$ is 3D coordinates and $\boldsymbol{n}_i^s \in \mathbb{R}^3$ is the unit normal vector. $\boldsymbol{\tau}_i^s \in \mathbb{R}^{\psi_{\tau}}$ and $\boldsymbol{\Upsilon}_i^s \in \mathbb{R}^{\psi_{\Upsilon}}$ capture its continuous and categorical physicochemical properties, such as hydrophobicity, hydrogen bonding, and charge~\citep{gainza2020deciphering,song2024surfpro}. 

This work designs peptides based on target proteins. Formally, given a peptide $C^{\text{pep}}$ with $n_\mathrm{pep}$ residues and a target protein $C^{\text{rec}}$ with $n_{\mathrm{rec}}$ residues, we aim to model the conditional joint distribution $p\left(C^{\mathrm{pep}} \mid C^{\mathrm{rec}}\right)$. The receptor can be sufficiently and succinctly parameterized as $C^{\text{rec}}=\left\{\left(a_i, O_i, \boldsymbol{x}_i, \boldsymbol{\chi}_i\right)\right\}_{i=1}^{n_{\mathrm{rec}}}$, where $\boldsymbol{\chi}_i[0]=\varphi_i$ and $\boldsymbol{\chi}_i \in[0,2 \pi)^5$. The ligand peptide surface is also portrayed, resulting in $C^{\mathrm{pep}}= \left\{\left(a_j, O_j, \boldsymbol{x}_j, \boldsymbol{\chi}_j\right)\right\}_{j=1}^{n_{\mathrm{pep}}} \cup \left\{\left(\boldsymbol{x}_i^s, \boldsymbol{n}_i^s, \boldsymbol{\tau}_i^s, \boldsymbol{\Upsilon}_i^s\right)\right\}_{i=1}^m$ with $m \gg n_{\mathrm{pep}}$. Practically, software like PyMol~\citep{delano2002pymol} or MSMS~\citep{robinson2014msm} can be utilized to compute the raw molecular surface of a protein.  


\paragraph{Probability Path and Flow.} Let $\mathbb{P}(\mathcal{M})$ be the probability distribution space on a manifold $\mathcal{M}$ with a Riemanian metric $g$. A probability path $p_t:[0,1]\times \mathcal{M} \rightarrow \mathbb{P}(\mathcal{M})$ is an interpolation between two distributions $p_0, p_1\in \mathbb{P}(\mathcal{M})$ indexed by a continuous parameter $t$. 

A flow on $\mathcal{M}$ is defined by a one-parameter diffeomorphism $\Phi: \mathcal{M} \rightarrow \mathcal{M}$, which is the result of integrating instantaneous deformations described by a time-varying vector field $u_t \in \mathcal{U}$. $u_t(x)\in\mathcal{T}_x \mathcal{M}$ is the gradient vector of the path $p_t$ on $x$ at time $t$.
By solving the following Ordinary Differential Equation (ODE) on $\mathcal{M}$ over $t\in [0,1]$ with an initial condition of $\phi_0(x)=x$: $\frac{\mathrm{d} \phi_t}{\mathrm{d}t} ({x}) = u_t(\phi_t(x)).$
We acquire the time-dependent flow $\phi_t: \mathcal{M} \rightarrow \mathcal{M}$ and the final diffeomorphism by setting $\Phi(x)=\phi_1(x)$. 
Given a source density $p_0$, $\phi_t(x)$ induces a push-forward operation $p_t=[\phi_t]_\# p_0$. It reshapes the point density $x\sim p_0$ to a more complicated one along $u_t(x)$, and the change-of-variable operator $\#$ is defined by $[\phi_t]_\# p_0(x) = p_0\left(\phi_t^{-1}(x)\right)\textrm{det}\left[\frac{\partial\phi_t^{-1}}{\partial x} (x)\right]$. The time-varying density $p_t$ is characterized by the Fokker-Planck equation: $\frac{\mathrm{d} p_t}{\mathrm{d} t} = -\textrm{div}(u_t p_t)$. Under these conditions,  $u_t$ is the probability flow for $p_t$, and $p_t$ is the marginal probability path generated by $u_t$.  
FM trains a Continuous Normalizing Flow (CNF) by fitting a vector field $v_\theta\in \mathcal{U}$ with parameters $\theta$ to a target vector field $u_t$ that produces a probability path $p_t$.  Its objective lies in the tangent space as:
\begin{equation}
    \mathcal{L}_{\textrm{RFM}}(\theta) = \mathbb{E}_{t\sim \mathcal{U}[0,1],x\sim p_t(x)} \|v_\theta(x, t) - u_t(x) \|^2_g, 
\end{equation}
As $u_t$ is intractable, we construct $p_t(x|x_1)$ with a conditional vector field $u_t(x|x_1)$. The objective becomes: $\mathcal{L}_{\textrm{CRFM}}(\theta)= \\ \mathbb{E}_{t\sim \mathcal{U}[0,1], x_1\sim p_1
(x_1), x\sim p_t(x|x_1)} \|v_\theta(x, t) - u_t(x|x_1) \|^2_g$, which shares the same gradients as $\mathcal{L}_{\textrm{RFM}}$~\citep{lipman2022flow}.  During inference, one can solve the ODE related to the neural vector field $v_\theta$ to push $x_0\in \mathcal{M}$ from $p_0$ to $p_1$ in time.

\paragraph{Continuous-Time Markov Chains.} CTMC~\citep{norris1998markov} is a class of discrete, continuous-time stochastic processes and is closely linked to probability flows. Suppose a categorical variable $x$ has $S$ states and its trajectory $x_t$ over time $t \in[0,1]$ follows a CTMC, $x_t$ alternates between resting in its current state and periodically jumping to another randomly chosen state. The frequency and destination of the jumps are determined by the rate matrix $R_t \in \mathbb{R}^{S \times S}$ with the constraint that its off-diagonal elements are non-negative. The probability $x_t$ will jump to a different state $j$ is $R_t\left(x_t, j\right) \mathrm{d} t$ for the next infinitesimal time step $\mathrm{d} t$. The transition probability is written as:
\begin{align}
\begin{split}
    p_{t+\mathrm{d} t \mid t}\left(j \mid x_t\right) & = \begin{cases}R_t\left(x_t, j\right) \mathrm{d} t & \text { for } j \neq x_t \\
    1+R_t\left(x_t, x_t\right) \mathrm{d} t & \text { for } j=x_t\end{cases} \\
    & =\delta\left\{x_t, j\right\}+R_t\left(x_t, j\right) \mathrm{d}t,
\end{split}
\end{align}
where $\delta\{i, j\}$ is the Kronecker delta. $\delta\{i, j\}$ is 1 when $i=j$ and is otherwise 0. $R_t\left(x_t, x_t\right):=-\sum_{k \neq x} R_t\left(x_t, k\right)$ in order for $p_{t+\mathrm{d} t \mid t}(\cdot \mid i)$ to sum to 1 . Using compact notation, $p_{t+\mathrm{d} t \mid t}$ is therefore a categorical distribution with probabilities $\delta\left\{x_t, \cdot\right\}+R_t\left(x_t, \cdot\right) \mathrm{d} t$ denoted as $\operatorname{Cat}\left(\delta\left\{x_t, j\right\}+R_t\left(x_t, j\right) \mathrm{d} t\right)$. Namely, $j \sim p_{t+\mathrm{d} t \mid t}\left(j \mid x_t\right) \Longleftrightarrow j \sim \operatorname{Cat}\left(\delta\left\{x_t, j\right\}+R_t\left(x_t, j\right) \mathrm{d} t\right)$. In practice, we need to simulate the sequence trajectory with finite time intervals $\Delta t$. A trajectory can be simulated with Euler steps~\citep{sun2022score}. 
\begin{equation}
    x_{t+\Delta t} \sim \operatorname{Cat}\left(\delta\left\{x_t, x_{t+\Delta t}\right\}+R_t\left(x_t, x_{t+\Delta t}\right) \Delta t\right),
\end{equation}
where the variable $x$ starts from an initial sample $x_0 \sim p_0$ at time $t=0$. The rate matrix $R_t$, along with an initial distribution $p_0$, together define the CTMC. 
With the marginal distribution at time $t$ as $p_t\left(x_t\right)$, the Kolmogorov equation allows us to relate the rate matrix $R_t$ to the change in $p_t\left(x_t\right)$:
\begin{equation}
    \partial_t p_t\left(x_t\right)=\underbrace{\sum_{j \neq x_t} R_t\left(j, x_t\right) p_t(j)}_{\text {incoming }}-\underbrace{\sum_{j \neq x_t} R_t\left(x_t, j\right) p_t\left(x_t\right)}_{\text {outgoing }}.
\end{equation}
The difference between the incoming and outgoing probability mass is the time derivative of the marginal $\partial_t p_t\left(x_t\right)$. Subsequently, we attain $\partial_t p_t=R_t^{\top} p_t$ where the marginals are treated as probability mass vectors: $p_t \in[0,1]^S$ and define an ODE in a vector space. The probability path $p_t$ is said to be generated by $R_t$ if $\partial_t p_t=R_t^{\top} p_t$ for $\forall t \in[0,1]$~\citep{campbell2024generative}.
\begin{figure*}[t]
    \centering
    \includegraphics[height=0.5\linewidth]{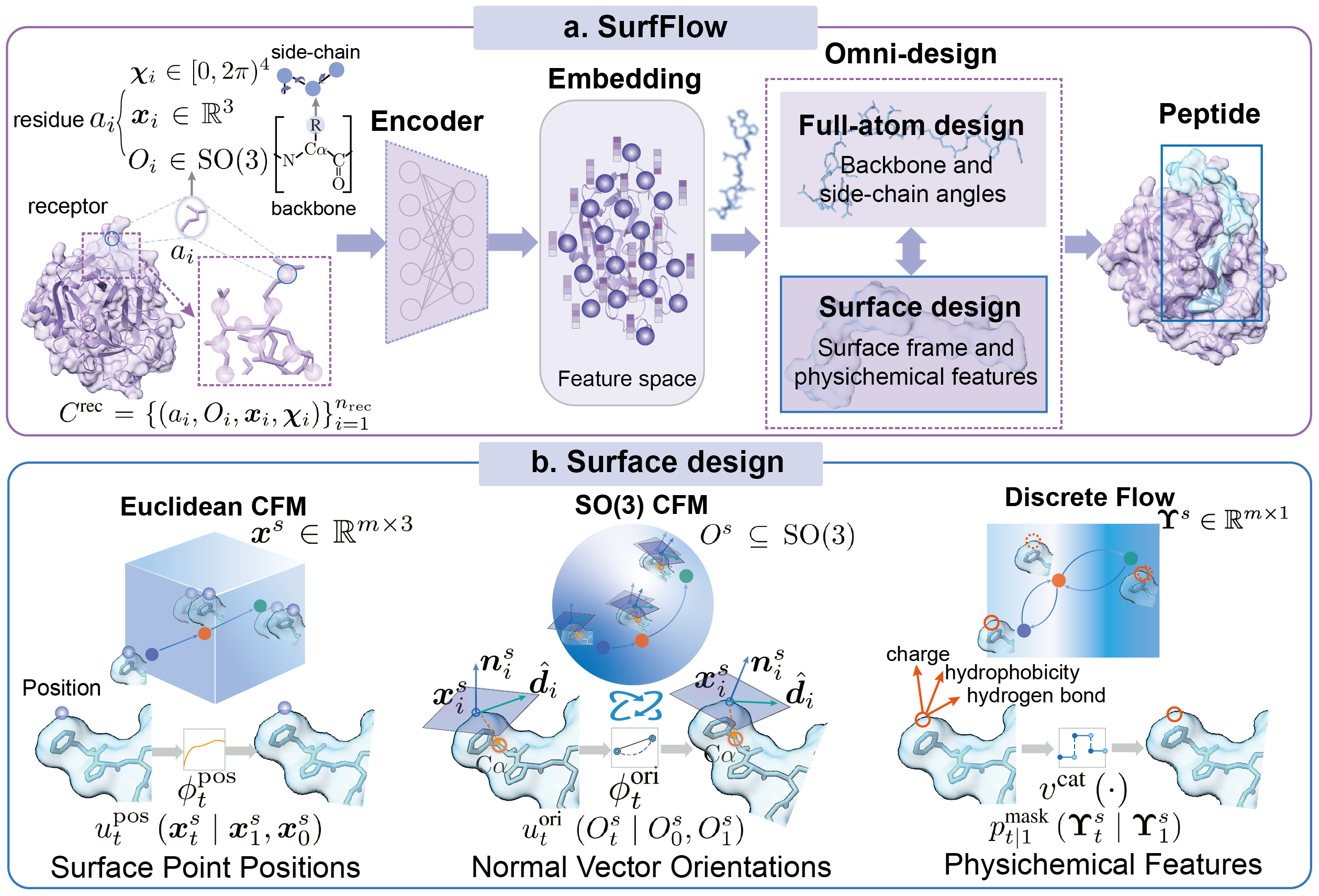}
    \caption{Workflow of SurfFlow for our peptide omni-design, which considers the multi-modality consistency among sequence, structure, and molecular surface during the generation process.}
    \Description{ }
    \label{fig:model}
\end{figure*}

\section{Method}
A molecular surface encapsulates both the 3D geometry of a protein in Euclidean space and its biochemical attributes, such as hydrophobicity and charge~\citep{gainza2020deciphering}. The interplay between surface shape and these biochemical properties is essential for defining a protein’s function. Given a target receptor with specific geometric and biochemical constraints, SurfFlow concurrently generates the peptide’s internal structure and external surface. Moreover, it can also account for key factors such as cyclicity and disulfide bonds (see Fig.~\ref{fig:model}).

\subsection{Flow Matching for Surface Generation}
A CFM framework is employed to learn the conditional peptide distribution based on its target protein $p\left(C^{\text{pep}} \mid C^{\text{rec}}\right)$. This joint probability is empirically decomposed into the product of probabilities of the internal structure elements and the external surface elements: 
\begin{equation}
\label{equ: propability}
\begin{split}
        p\left(C^{\mathrm{pep}} \mid C^{\mathrm{rec}}\right) \propto p\left(\left\{\left(a_j, O_j, \boldsymbol{x}_j, \boldsymbol{\chi}_j\right)\right\}_{j=1}^{n_{\mathrm{pep}}} \big| C^{\mathrm{rec}}\right) \\ p\left(\left\{\left(\boldsymbol{x}_i^s, \boldsymbol{n}_i^s,  \boldsymbol{\tau}_i^s, \boldsymbol{\Upsilon}_i^s\right)\right\}_{i=1}^m \big|  C^{\mathrm{rec}}\right), 
\end{split}
\end{equation}
where $p\left(\left\{\left(\boldsymbol{x}_i^s, \boldsymbol{n}_i^s,  \boldsymbol{\tau}_i^s, \boldsymbol{\Upsilon}_i^s\right)\right\}_{i=1}^m \big|  C^{\mathrm{rec}}\right)$ is further separated as the product of four basic elements $p\left(\left\{\boldsymbol{x}_i^s\right\}_{i=1}^m \big|  C^{\mathrm{rec}}\right)$, $p\left(\left\{\boldsymbol{n}_i^s\right\}_{i=1}^m \big|  C^{\mathrm{rec}}\right)$, $\\ p\left(\left\{\boldsymbol{\tau}_i^s\right\}_{i=1}^m \big|  C^{\mathrm{rec}}\right)$, and $p\left(\left\{\boldsymbol{\Upsilon}_i^s\right\}_{i=1}^m \big|  C^{\mathrm{rec}}\right)$. The construction of different probability flows on the surface point's position $p\left(\boldsymbol{x}_i^s \mid C^{\text{rec}}\right)$, orientation $p\left(\boldsymbol{n}_i^s \mid C^{\text{rec}}\right)$, continuous properties $p\left(\boldsymbol{\tau}_i^s \mid C^{\text{rec}}\right)$, and categorical properties $p\left(\boldsymbol{\Upsilon}_i^s \mid C^{\text{rec}}\right)$ is elaborated as follows.  

\paragraph{Physichemical Features.} FM is typically applied to continuous spaces. However, certain biochemical characteristics take on discrete and categorical values. For example, each point can be classified based on its hydrogen bonding potential: donor, acceptor, or neutral. This challenge also arises in protein generation tasks that focus solely on structure~\citep{luo2022antigen}, where residue types follow a categorical distribution. To address this, previous studies~\citep{li2024full,lin2024ppflow} employ soft one-hot encoding to map categorical distributions to a probability simplex or directly apply FM to multinomial distributions. However, this straightforward approach may result in suboptimal performance for protein co-design. Recently, more advanced FM methods tailored for discrete spaces have been proposed~\citep{campbell2024generative,gat2024discrete,stark2024dirichlet} to overcome these limitations.

Discrete flow models (DFMs) are generative algorithms designed to operate in discrete spaces by simulating a probability flow that transitions from noise to data. DFMs trace a trajectory of discrete variables that align the noise-to-data flow, allowing the generation of new samples. Building on the work of~\citet{campbell2024generative, gat2024discrete}, we implement a DFM using CTMCs for the biochemical properties $\boldsymbol{\Upsilon}^s \in \mathbb{R}^{m \times \psi_{\Upsilon}}$. Specifically, we train a neural network $v^{\text{cat}}\left(\cdot\right)$ to approximate the true denoising distribution $p_{1\mid t}\left(\boldsymbol{\Upsilon}^s_t \mid \boldsymbol{\Upsilon}^s_1\right)$. This is done using a cross-entropy loss, where the model learns to predict the clean data point $\boldsymbol{\Upsilon}^s_1$ when given a noisy input $\boldsymbol{\Upsilon}^s_t \sim p_{t \mid 1}\left(\boldsymbol{\Upsilon}^s_t \mid \boldsymbol{\Upsilon}^s_1\right)$ as:
\begin{equation}
\label{equ: loss_discrete}
    \mathcal{L}_{\mathrm{cat}}(\theta)=\mathbb{E}_{t\sim \mathcal{U}(0,1), p\left(\boldsymbol{\Upsilon}^s_1\right), p_{t \mid 1}\left(\boldsymbol{\Upsilon}^s_t \mid \boldsymbol{\Upsilon}^s_1\right)}\left[\log v^{\text{cat}}\left(\boldsymbol{\Upsilon}^s_t,t, C^{\mathrm{rec}}\right)\right].  
\end{equation}
Here, rather than a linear interpolation towards $\boldsymbol{\Upsilon}^s_1$ from a uniform prior $p_{t\mid 1}^{\text{unif}}\left(\boldsymbol{\Upsilon}^s_t \mid \boldsymbol{\Upsilon}^s_1\right) = \operatorname{Cat}\left(t\delta\{\boldsymbol{\Upsilon}^s_1,\boldsymbol{\Upsilon}^s_t\} + (1-t) \frac{1}{S}\right)$, we adopt an artificially introduced mask state $\text{M}$ and the conditional path becomes~\citep{campbell2024generative}:
\begin{equation}
\label{equ: p_t_1}
    p_{t \mid 1}^{\text{mask}}\left(\boldsymbol{\Upsilon}^s_t \mid \boldsymbol{\Upsilon}^s_1\right) = \text{Cat}(t\delta\{\boldsymbol{\Upsilon}^s_1,\boldsymbol{\Upsilon}^s_t\} + (1-t)\delta \{\text{M}, \boldsymbol{\Upsilon}^s_t\}).
\end{equation}
Notably, $\mathcal{L}^{\mathrm{cat}}(\cdot)$ has a strong relation to the model loglikelihood and the Evidence Lower Bound (ELBO) used to train diffusion models~\citep{campbell2024generative}. It also does not depend on $R_t\left(\boldsymbol{\Upsilon}^s_t, j \mid \boldsymbol{\Upsilon}^s_1\right)$. 
There are many choices for $R_t\left(\boldsymbol{\Upsilon}^s_t, j \mid \boldsymbol{\Upsilon}^s_1\right)$ that all generate the same $p_{t \mid 1}\left(\boldsymbol{\Upsilon}^s_t \mid \boldsymbol{\Upsilon}^s_1\right)$. At inference time, we pick the rate matrix for $\boldsymbol{\Upsilon}^s_t \neq j$ as:
\begin{equation}
  R_t^*\left(\boldsymbol{\Upsilon}^s_t, j \mid \boldsymbol{\Upsilon}^s_1\right):=\frac{\operatorname{ReLU}\left(\partial_t p_{t \mid 1}\left(j \mid \boldsymbol{\Upsilon}^s_1\right)-\partial_t p_{t \mid 1}\left(\boldsymbol{\Upsilon}^s_t \mid \boldsymbol{\Upsilon}^s_1\right)\right)}{S \cdot p_{t \mid 1}\left(\boldsymbol{\Upsilon}^s_t \mid \boldsymbol{\Upsilon}^s_1\right)},  
\end{equation}
where $\operatorname{ReLU}(a)=\max (a, 0)$ and $\partial_t p_{t \mid 1}$ can be found by differentiating our explicit form for $p_{t \mid 1}$ in Equ.~\ref{equ: p_t_1}. This choice of $R_t^*$ assumes $p_{t \mid 1}\left(\boldsymbol{\Upsilon}^s_t \mid \boldsymbol{\Upsilon}^s_1\right)>0$. 

\paragraph{Position.} Euclidean CFM is utilized to generate surface point positions $\boldsymbol{x}^s \in \mathbb{R}^{m\times 3}$. We adopt the standard isotropic Gaussian $\mathcal{N}\left(0, \boldsymbol{I}_3\right)$ as the prior, with the target distribution being $p\left(\boldsymbol{x}^s \mid C^{\text{rec}}\right)$. The conditional flow is defined as a linear interpolation between sampled noise $\boldsymbol{x}_0^s \sim \mathcal{N}\left(0, \boldsymbol{I}_3\right)$ and data points $\boldsymbol{x}_1^s \sim p\left(\boldsymbol{x}^s \mid C^{\text{rec}}\right)$. This linear interpolation ensures a straight trajectory, contributing to training and sampling efficiency by following the shortest path between two points in Euclidean space~\citep{liu2022flow}. The conditional vector field $u_t^{\text{pos}}$ is obtained by taking the time derivative of the linear flow $\phi_t^{\text{pos}}$ using Independent Coupling techniques:
\begin{align}
    \phi_t^{\mathrm{pos}}\left(\boldsymbol{x}_0^s, \boldsymbol{x}_1^s\right)&=t \boldsymbol{x}_1^s+(1-t) \boldsymbol{x}_0^s, \\
    u_t^{\mathrm{pos}}\left(\boldsymbol{x}_t^s \mid \boldsymbol{x}_1^s, \boldsymbol{x}_0^s\right)&=\boldsymbol{x}_1^s-\boldsymbol{x}_0^s=\frac{\boldsymbol{x}_1^s-\boldsymbol{x}_t^s}{1-t}.
\end{align}
We use a time-dependent translation-invariant surface network $v^{\text{pos}}(\cdot)$ to predict the conditional vector field based on the current interpolant $\boldsymbol{x}_t$ and the timestep $t$. The CFM objective of the surface point cloud position is formulated as:
\begin{equation}
\begin{split}
\label{equ: loss_pos}
    \mathcal{L}_{\mathrm{pos}}(\theta)=\mathbb{E}_{t\sim \mathcal{U}(0,1), p\left(\boldsymbol{x}_1^s\right), p\left(\boldsymbol{x}_0^s\right), p\left(\boldsymbol{x}_t^s|\boldsymbol{x}_0^s,\boldsymbol{x}_1^s\right)}\\
    \left\|v^{\mathrm{pos}}\left(\boldsymbol{x}_t^s, t, C^{\mathrm{rec}}\right)-\left(\boldsymbol{x}_1^s-\boldsymbol{x}_0^s\right)\right\|_2^2,
\end{split}
\end{equation}
where $\mathcal{U}(0,1)$ is a uniform distribution on $[0,1]$. During generation, we first sample from the prior $\boldsymbol{x}_0^s \sim$ $\mathcal{N}\left(0, \boldsymbol{I}_3\right)$ and solve the probability flow with the learned predictor $v^{\text{pos}}(\cdot)$ using the $N$-step forward Euler method to get the position of residue $j$ with $t=\left\{0, \ldots, \frac{N-1}{N}\right\}$ :
\begin{equation}
    \boldsymbol{x}_{t+\frac{1}{N}}^s=\boldsymbol{x}_t^s+\frac{1}{N} v^{\mathrm{pos}}\left(\boldsymbol{x}_t^s, t, C^{\mathrm{rec}}\right).
\end{equation}

\paragraph{Normal Vector Orientations.} The normal vector $\boldsymbol{n}_i^s$ is a unit vector perpendicular to the tangent plane of the surface at $\boldsymbol{x}_i^s$. It reveals essential geometric information of the surface orientation and curvature, directly tied to the protein's functional~\citep{song2024surfpro}. Convex regions, for example, might be more accessible for binding, while concave regions might be better suited for pockets or clefts involved in substrate binding~\citep{laskowski1996protein}. To capture its orientation, we construct a set of rotation matrices $O^s \subseteq \mathrm{SO}(3)$ concerning the global frame, whose element is defined by $O^s_i = \left(\boldsymbol{n}_i^s, \hat{\boldsymbol{d}}_i, \boldsymbol{n}_i^s \times  \hat{\boldsymbol{d}}_i \right)\in\mathrm{SO}(3)$. Here, $\hat{\boldsymbol{d}}_i$ is a unit vector orthogonal to $\boldsymbol{n}_i^s$ and is acquired by normalizing the cross product between $\boldsymbol{n}_i^s$ and the direction pointing from the surface point $\boldsymbol{x}_i^s$ to its nearest $\text{C}_\alpha$ coordinate. This frame $O^s_i$ effectively describes the geometric relationship between the protein surface and the underlying backbone structure.

The 3D rotation group $\mathrm{SO}(3)$ is a smooth Riemannian manifold, with its tangent space, $\mathfrak{so}(3)$, forming a Lie algebra consisting of skew-symmetric matrices. Elements of $\mathfrak{so}(3)$ can be viewed as infinitesimal rotations around specific axes and represented as rotation vectors in $\mathbb{R}^3$~\citep{blanco2021tutorial}. We adopt the uniform distribution over $\mathrm{SO}(3)$ as the prior $p\left(O_s^0\right)$. Just as FM in Euclidean space is based on the shortest path between two points, we extend this idea to $\mathrm{SO}(3)$ by using geodesics, which define the minimal rotational distance between two orientations~\citep{lee2018introduction}. These geodesics provide a natural framework for interpolating and evolving rotations while respecting the geometry of the manifold~\citep{bose2023se,yim2023fast}. The conditional flow $\phi^{\text{ori}}$ and vector field $u_t^{\text{ori}}$ are constructed by geodesic interpolation between $O^s_0 \subseteq \mathcal{U}(\mathrm{SO}(3))$ and $O^s_1 \in p\left(O^s \mid C^{\text{rec}}\right)$, with the geodesic distance decreasing linearly over time:
\begin{align}
    \phi_t^{\text {ori }}\left(O^s_0, O^s_1\right) & =\exp_{O^s_0}\left(t \log _{O^s_0}\left(O^s_1\right)\right), \\
    u_t^{\text {ori }}\left(O^s_t \mid O^s_0, O^s_1\right) & =\frac{\log_{O^s_t} \left(O^s_1\right)}{1-t},
\end{align}
where $\text{exp}(\cdot)$ and $\text{log}(\cdot)$ are the exponential and logarithm maps on $\mathrm{SO}(3)$ that can be computed efficiently using Rodrigues' formula. A rotation-equivariant surface network $v^{\text {ori}}(\cdot)$ is applied to predict the vector field $u_t^{\text{ori}}$, represented as rotation vectors. The CFM objective on $\mathrm{SO}(3)$ is formulated as:
\begin{equation}
\begin{split}
\label{equ: loss_ori}
    \mathcal{L}_{\text{ori}}(\theta)=\mathbb{E}_{t\sim \mathcal{U}(0,1), p\left(O^s_1\right), p\left(O^s_0\right),p\left(O^s_t|O^s_0,O^s_1\right)}\\
    \left\|v^{\text{ori}}\left(O^s_t, t, C^{\mathrm{rec}}\right)-\frac{\log _{O^s_t} \left(O^s_1\right)}{1-t}\right\|_{\mathrm{SO}(3)}^2,
\end{split}
\end{equation}
where the vector field $v^{\text{ori}}(\cdot)$ is defined in the tangent space $\mathfrak{so}(3)$ of $\mathrm{SO}(3)$, with the norm $|\cdot |^2$ derived from the canonical metric on $\mathrm{SO}(3)$. In the inference phase, the process is initialized at $O^s_0 \sim \mathcal{U}(\mathrm{SO}(3))$ and proceeds by following the geodesic in $\mathrm{SO}(3)$, taking small steps over time $t$:
\begin{equation}
    O_{t+\frac{1}{N}}=\exp _{O^s_t}\left(\frac{1}{N} v^{\text {ori }}\left(O^s_t, t, C^{\mathrm{rec}}\right)\right). 
\end{equation}
\paragraph{Overall Training Loss.} Combining all modalities, the final FM objective is for conditional peptide generation obtained as the weighted sum of three loss functions in Equ.~\ref{equ: loss_pos}, ~\ref{equ: loss_ori}, and~\ref{equ: loss_discrete} as well as several additional losses. It can be written as:
\begin{equation}
\label{equ:overall_loss}
    \mathcal{L}_{\mathrm{CFM}} = \lambda_{\mathrm{pos}} \mathcal{L}_{\mathrm{pos}} + \lambda_{\mathrm{ori}} \mathcal{L}_{\mathrm{ori}} + \lambda_{\mathrm{cat}} \mathcal{L}_{\mathrm{cat}} + \lambda_{\mathrm{con}} \mathcal{L}_{\mathrm{con}} + \lambda_{\mathrm{str}}  \mathcal{L}_{\mathrm{str}}, 
\end{equation}
where $\lambda_*$ are the hyperparameters to control the impact of different loss components. $\mathcal{L}_{\mathrm{con}}$ is the loss function for continuous biochemical properties, and $\mathcal{L}_{\mathrm{str}}$ is the FM objective for modeling the factorized distribution of residues' positions $\left\{a_j\right\}_{j=1}^{n_{\mathrm{pep}}}$, orientations $\left\{O_j\right\}_{j=1}^{n_{\mathrm{pep}}}$, amino acid types $\left\{\boldsymbol{x}_j\right\}_{j=1}^{n_{\mathrm{pep}}}$, and side-chain torsion angles $\left\{\boldsymbol{\chi}_j\right\}_{j=1}^{n_{\mathrm{pep}}}$ as discussed in Equ.~\ref{equ: propability} and App.~\ref{app:fm_internal}. The network details to parameterize the generation procedure are illustrated in App.~\ref{app:networks}. 

\subsection{Equivariant Surface Geometric Network}
The success of SurfFlow is intricately tied to the careful design of the backbone architectures for $v^{\text{cat}}(\cdot)$, $v^{\text{pos}}(\cdot)$, and $v^{\text{ori}}(\cdot)$, which, however, face several fundamental challenges. The first stems from the irregular geometry of molecular surfaces, which are inherently continuous, highly intricate, and exhibit complex features. This makes it difficult to accurately capture their structure using conventional discrete grids or graph-based representations. Second, scalability presents a major hurdle, as molecular surfaces can contain millions of vertices, leading to computational bottlenecks in processing such large-scale data. Recent advancements in equivariant geometric encoders~\citep{sverrisson2021fast, gainza2020deciphering,somnath2021multi} have shown promise in modeling molecular surfaces with strong performance on predictive tasks, such as binding affinity prediction~\citep{wu2024surface}, ligand binding site prediction~\citep{zhang2023equipocket}, and inverse folding~\citep{tang2024bc}. However, their applicability to generative tasks, particularly those requiring the co-design of proteins, remains largely unexplored. More critically, these approaches do not account for the intricate inter-protein interactions that are central to tasks like protein-peptide design, leaving a significant gap in their utility for target-aware generative modeling.

To bridge this gap, we propose an equivariant surface geometric network (ESGN) specifically tailored for our protein design challenge with several key innovations. Notably, for the receptor pocket, we can also generate its surface point cloud $Q^{\text{rec}} = \{q_j^{\text{rec}}\}_{j=1}^{m_{\text{rec}}}$, each entity with associated static attributes $\left(\boldsymbol{x}_j^{s,\text{rec}}, \boldsymbol{n}_j^{s,\text{rec}}, \boldsymbol{\tau}_j^{s,\text{rec}}, \boldsymbol{\Upsilon}_j^{s,\text{rec}}\right)$. Consequently, we can build a dynamic heterogeneous surface graph $\mathcal{G} = (\mathcal{G}_{\text{rec}}, \mathcal{G}_{\text{pep}})$ at different timestep $t$, where the \emph{intra} and \emph{inter}-point connectivities $(\mathcal{E}_{\text{rec}}, \mathcal{E}_{\text{pep}}, \mathcal{E}_{\text{inter}})$ are determined based on a spatial distance threshold cutoff $r$. We initialize the node feature of each graph by a simple mapping on their SE(3)-invariant physicochemical properties, obtaining $\boldsymbol{h}_i^{(0)} = f_e(\boldsymbol{\Upsilon}_i^s)$ and $\boldsymbol{h}_j^{(0),\text{rec}} = f_e(\boldsymbol{\Upsilon}_j^{s,\text{rec}})$, where $f_e:\mathbb{R}^{\psi_{\Upsilon}} \rightarrow \mathbb{R}^{\psi_{h}}$ is a multi-layer perception (MLP). 

Then before forwarding $\boldsymbol{h}_i^{(0)}$ and $\boldsymbol{h}_j^{(0),\text{rec}}$ to each layer $l\in[L]$ of our ESGN, we compute crucial scalars to describe the local geometries: one is the inter-point distance $d_{ij}^s = \| \boldsymbol{x}_{i}^s - \boldsymbol{x}_{j}^s\|^2$ and $d_{ij}^{s,\text{rec}} =\| \boldsymbol{x}_{i}^{s,\text{rec}} - \boldsymbol{x}_{i}^{s,\text{rec}}\|^2$, and the other is the angles between normals and the connecting directed line of two points $(\boldsymbol{x}_{ij}, \boldsymbol{x}_{ji})$ by $\varphi_{\mathbf{n}_i ij}=\angle \boldsymbol{n}_i \boldsymbol{x}_{ij}$ and $\varphi_{\mathbf{n}_j ji}=\angle \boldsymbol{n}_{j} \boldsymbol{x}_{ji}$, where two intersecting planes can be formulated with normals of two surface points. After that, the \emph{intra}-graph messages for any inter-connected nodes (\emph{i.e.}, $\forall e_{j\rightarrow i}\in \mathcal{E}_{\text{rec}} \cup \mathcal{E}_{\text{pep}}$) are updated as~\footnote{The superscript \emph{rec} is omitted for conciseness.}:
\begin{align}
    \boldsymbol{m}_{j\rightarrow i} &= f_m\left(\boldsymbol{h}_i^{(l)}, \boldsymbol{h}_j^{(l)}, \boldsymbol{e}_{\text{RBF}}^{(ij)}, \boldsymbol{a}_{\text{SBF}}^{(\mathbf{n}_i ij)}, \boldsymbol{a}_{\text{SBF}}^{(\mathbf{n}_j ji)} \right), \\ 
    \boldsymbol{m}_{j\rightarrow i}' &=\varpi_{ij} \cdot \boldsymbol{m}_{j\rightarrow i},\quad \varpi_{ij}=\frac{\exp(\boldsymbol{W}_{\mathrm{m}}\boldsymbol{m}_{j\rightarrow i} + {b}_{\mathrm{m}})}{\sum_{k\in\mathcal{N}_{(i)}}\exp(\boldsymbol{W}_{\mathrm{m}}\boldsymbol{m}_{k\rightarrow i} + {b}_{\mathrm{m}})},
\end{align}
where $f_m(\cdot)$ is another MLP appended to transform each pair of node features into messages $\boldsymbol{m}_{j\rightarrow i}\in\mathbb{R}^{\psi_m}$. $\boldsymbol{e}^{(ij)}_{\text{RBF}}$ is the radial basis function representation of $d_{ij}^{s} $ and $d_{ij}^{s,\text{rec}} $. $\boldsymbol{a}_{\mathrm{SBF}}^{(\mathbf{n}_i ij)}$ and $\boldsymbol{a}_{\mathrm{SBF}}^{(\mathbf{n}_j ji)}$ are 3D spherical Fourier-Bessel representations~\citep{gasteiger2020fast}. Then a softmax is employed to reweight the messages, where the weight matrix $\boldsymbol{W}_{\mathrm{m}}\in\mathbb{R}^{\psi_{m}\times 1}$ and the vector ${b}_{\mathrm{m}}\in\mathbb{R}$ are learnable and $\mathcal{N}_{(i)}^{\text{intra}} = \left\{j| e_{j\rightarrow i}\in \mathcal{E}_{\text{rec}} \cup \mathcal{E}_{\text{pep}} \right\}$ is the neighborhood set of node $i$ within the same protein. 

In addition to intra-surface interactions, we attain cross-graph messages of inter-connections for $\forall e_{j\rightarrow i}\in \mathcal{E}_{\text{inter}}$, using a distance-weighted attention technique~\citep{ganea2021independent,wu2022pre}:
\begin{align}
    a_{j \rightarrow i} &=\frac{\exp \left(\left\langle f_q\left(\mathbf{h}_{i}^{(t), l}\right), f_k\left(\mathbf{h}_{j}^{(l)}\right)\right\rangle\right)}{\sum_{j^{\prime}} \exp \left(\left\langle f_q\left(\mathbf{h}_{i}^{(t), l}\right), f_k\left(\mathbf{h}_{j^{\prime}}^{(t), l}\right)\right\rangle\right)}, \\
    \boldsymbol{\mu}_{j \rightarrow i} &=a_{j \rightarrow i} \mathbf{h}_{j}^{(l)} \cdot \phi_d\left({d_{ij}^{s}}^{(l)}\right), \forall e_{j\rightarrow i} \in \mathcal{E}_{\text{inter}}, 
\end{align}
where $a_{j \rightarrow i}$ is an attention score calculated by trainable MLPs $f_{q}(\cdot)$ and $f_{k}(\cdot)$. $\phi_d(\cdot)$ operates on the inter-point distances ${d_{ij}^s}^{(l)}$ to acquire $\boldsymbol{\mu}_{j \rightarrow i}\in\mathbb{R}^{\psi_\mu}$.  After that, both \emph{intra}- and \emph{inter}-messages are aggregated and propagated from the vicinity of each point $q_i$ to update its node feature and coordinates as: 
\begin{align}
    \boldsymbol{h}_{i}^{(l+1)} &= f_{h}\left(\boldsymbol{h}_{i}^{(l)}, \sum_{j\in \mathcal{N}_{(i)}^{\text{intra}} }\boldsymbol{m}_{j\rightarrow i}, \sum_{j'\in \mathcal{N}_{(i)}^{\text{inter}}} \boldsymbol{\mu}_{j'\rightarrow i}\right), \\
    \quad \boldsymbol{x}_{i}^{(l+1)} &=\boldsymbol{x}_{i}^{(l)}+\frac{1}{\|\mathcal{N}_{(i)}^{\text{intra}}\|} \sum_{j\in \mathcal{N}_{(i)}^{\text{intra}}} f_{x}^{\text{intra}} \left(\boldsymbol{m}_{j\rightarrow i} \right) \boldsymbol{x}_{ij}^{(l)} \\
    &+ \frac{1}{\|\mathcal{N}_{(i)}^{\text{inter}}\|} \sum_{j'\in \mathcal{N}_{(i)}^{\text{inter}}} f_{x}^{\text{inter}}\left(\boldsymbol{\mu}_{j'\rightarrow i} \right) \boldsymbol{x}_{ij'}^{(l)},   
\end{align}
where $f_{h}(\cdot)$ is the node operation implemented as an MLP to aggregate the \emph{intra}- and cross-graph messages. $f_x^{\text{intra}}:\mathbb{R}^{\psi_m}\rightarrow \mathbb{R}$ and $f_x^{\text{inter}}:\mathbb{R}^{\psi_\mu}\rightarrow \mathbb{R}$ transform $\boldsymbol{m}_{j\rightarrow i}$ and $\boldsymbol{\mu}_{j'\rightarrow i}$ into scalar scores, respectively, to control the impact of directional vectors $\boldsymbol{x}_{ij}^{(l)}=\boldsymbol{x}_{i}^{(l)} - \boldsymbol{x}_{j}^{(l)}$ and $\boldsymbol{x}_{ij'}^{(l)}=\boldsymbol{x}_{i}^{(l)} - \boldsymbol{x}_{j'}^{(l)}$. $\mathcal{N}_{(i)}^{\text{inter}} = \left\{j| e_{j\rightarrow i}\in \mathcal{E}_{\text{inter}} \right\}$ is the neighborhood set of node $i$ in the counterpart protein. Remarkably, as the flow interpolates the coordinates of surface point clouds $\boldsymbol{x}^s_t$ at different timesteps $t\in[0,1]$, \emph{inter-} and \emph{intra}-connectivities $(\mathcal{E}_{\text{inter}}, \mathcal{E}_{\text{pep}})$ are dynamically changing and have to be rebuilt according to the distance threshold $r$. It can be easily concluded that ESGN preserves the equivariance. 

\begin{table*}[t]
    \caption{Evaluation of different methods in the sequence-structure co-design task and ablation studies on key components of SurfFlow. The \textbf{best} and \underline{suboptimal} results are labeled boldly and underlined. }
    \centering
    \resizebox{1.85\columnwidth}{!}{%
    \begin{tabular}{ccccccccc} \toprule
    & \multicolumn{4}{c}{ \textbf{Geometry} } & \multicolumn{2}{c}{ \textbf{Energy} } & \multicolumn{2}{c}{ \textbf{Design} } \\
    & AAR $\% \uparrow$ & RMSD $\text{\AA} \downarrow$ & $\operatorname{SSR} \% \uparrow$ & BSR $\% \uparrow$ & Stb. $\% \uparrow$ & Aff. $\% \uparrow$ & Des. $\% \uparrow$ & Div. $\uparrow$ \\ \midrule
    RFdiffusion~\citep{watson2023novo} & 40.14 & 4.17 & 63.86 & {26.71} & \textbf{26.82} & 16.53 & \textbf{78.52} & 0.38 \\
    ProteinGen~\citep{lisanza2023joint} & 45.82 & 4.35 & 29.15 & 24.62 & 23.48 & 13.47 & 71.82 & 0.54 \\
    Diffusion~\citep{luo2022antigen} & 47.04 & 3.28 & 74.89 & 49.83 & 15.34 & 17.13 & 48.54 & \underline{0.57} \\
    PepGLAD~\citep{kong2024full}  & 50.43 & 3.83 & 80.24 & 19.34 & 20.39 & 10.47 & \underline{75.07} & 0.32  \\
    PPIFlow~\citep{lin2024ppflow} & 48.35  &  3.59 & 68.13 & 25.94 & 15.77 & 12.08 & 46.53 & 0.51 \\ 
    PepFlow~\citep{li2024full} & \underline{51.25} & \underline{2.07} & \underline{83.46} & \underline{86.89} & 18.15 & 21.37 & 65.22 & 0.42 \\ \midrule
    \grow{SurfFlow (w/o ESGN)} & 52.59 & 2.05 & 83.77 & 86.91 & 19.42 & 19.82 & 68.41 & 0.60 \\ 
    \grow{SurfFlow (w/o Position)} & 53.26 & 1.99 & 84.79 & 87.15 & 21.30 & 22.38 & 72.09 & 0.60 \\ 
    \grow{SurfFlow (w/o Orientation)} & 53.04 & 2.00 & 84.60 & 87.04 & 20.79 & 22.46 & 72.36 & 0.60 \\ 
    \grow{SurfFlow (w/o Biophysical Prop.)} & 52.31 & 2.03 & 83.96 & 86.98 & 19.55 & 20.25 & 70.83 & 0.58\\ 
    \grow{\textbf{SurfFlow}} & \textbf{54.07} & \textbf{1.96} & \textbf{85.11} & \textbf{87.38} & \underline{22.46} & \textbf{22.51} & {73.60} & \textbf{0.61} \\ \bottomrule
    \end{tabular}}
    \label{tab:co-design}
\end{table*}
\begin{figure*}[t]
    \centering
    \includegraphics[width=1.0\linewidth]{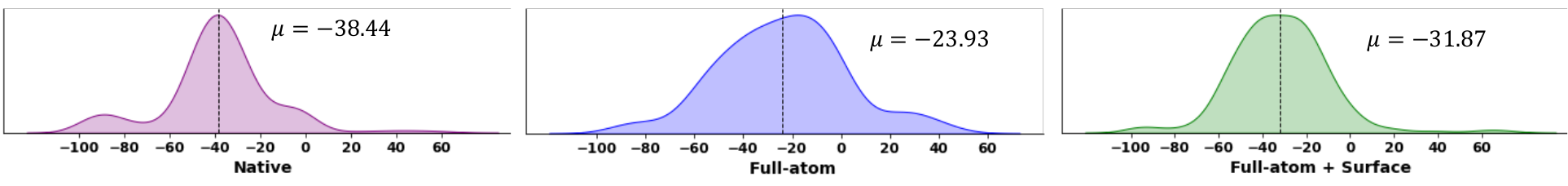}
    \vspace{-1em}
    \caption{Binding energy distributions of designed and native peptides, where the lower is better. }
    \Description{ }
    \label{fig:energy}
\end{figure*}

\subsection{Condtional Protein Design}
Inspired by the success of controllable image generation~\citep{zhang2023adding}, we propose peptide design conditioned on key factors $c$, such as sequence length $n_{\mathrm{pep}}$, cyclicity, and the presence of disulfide bonds. Our objective becomes $p\left(C^{\text{pep}} \mid C^{\text{rec}}, c\right)$, allowing flow models to incorporate additional conditions. Accordingly, the vector field networks are adapted to $v^{\mathrm{pos}}\left(\boldsymbol{x}_t^s, t, C^{\mathrm{rec}}, c\right)$, $v^{\text {ori }}\left(O^s_t, t, C^{\mathrm{rec}}, c\right)$, and $v^{\text{cat}}\left(\boldsymbol{\Upsilon}^s_t, t, C^{\mathrm{rec}}, c\right)$. In practice, we focus on two primary goals: (1) \emph{\textbf{Cyclic peptides}} offer enhanced stability by constraining the backbone, thus reducing conformational flexibility and increasing resistance to enzymatic degradation. This structure form improves binding affinity due to more defined and stable conformations. (2) \emph{\textbf{Disulfide bonds}}, covalent interactions between cysteine residues, assist in proper folding and structural stabilization. These bonds protect peptides from oxidative damage and proteolytic enzymes, enhancing their resistance to harsh industrial conditions. Additionally, disulfide bonds can improve biological activity by creating conformational constraints, further enhancing the therapeutic potential of peptide-based drugs.

\section{Experiments}
We comprehensively evaluate SurfFlow on unconditioned and conditioned sequence-structure co-design tasks and the side-chain packing problem. For benchmarking, we use the PepMerge dataset derived from PepBDB~\citep{wen2019pepbdb} and Q-BioLip~\citep{wei2024q}. Following~\citet{li2024full}, we cluster complexes based on 40\% sequence identity using mmseqs2~\citep{steinegger2017mmseqs2}, after filtering out duplicates and applying empirical criteria (\emph{e.g.}, resolution $<$ 4\AA, peptide length between 3 and 25). This yields 8,365 non-redundant complexes across 292 clusters. The test set consists of 10 clusters and 158 complexes. More details and additional results are elaborated on App.~\ref{app: exp_details}.

\subsection{Unconditioned Sequence-structure Co-design}
\paragraph{Baselines.} Two lines of state-of-the-art protein design approaches are chosen as baselines. The first kind ignores the side-chain conformations, including RFDiffusion~\citep{watson2023novo} and ProteinGen~\citep{lisanza2023joint}. RFDiffusion produces protein backbones, and sequences are later forecast by ProteinMPNN~\citep{dauparas2022robust}. ProteinGen improves RFDiffusion by jointly sampling backbones and corresponding sequences. 
The other kind considers a full-atom style protein generation, including Diffusion~\citep{luo2022antigen}, PepGLAD~\citep{kong2024full}, and PepFlow~\citep{li2024full}. 

\paragraph{Evaluation Metrics.} Generated peptides are evaluated from three key aspects. (1) \textbf{Geometry}: Designed peptides should closely resemble native sequences and structures. We use the amino acid recovery rate (\textbf{AAR}) to quantify sequence identity between generated peptides and ground truth. Structural similarity is assessed through the root-mean-square deviation (\textbf{RMSD}) of $C_\alpha$ atoms after aligning the complexes. Secondary-structure similarity ratio (\textbf{SSR}) measures the proportion of shared secondary structures, while the binding site ratio (\textbf{BSR}) compares the overlap between the binding sites of the generated and native peptides on the target protein. 
(2) \textbf{Energy}: Our goal is to design high-affinity peptide binders that enhance the stability of protein-peptide complexes. \textbf{Affinity} is defined as the percentage of generated peptides with higher binding affinities (lower binding energies) than the native peptide, while \textbf{Stability} indicates the proportion of complexes with lower total energy than the native state. Energy calculations are performed using Rosetta~\citep{alford2017rosetta}. (3) \textbf{Design}: \textbf{Designability} reflects the consistency between designed sequences and structures. It is measured by the fraction of sequences that can fold into structures similar to their corresponding generated forms, with $C_\alpha$ RMSD $<$ 2~\textup{\AA} as the threshold. We use ESMFold~\citep{lin2022language} to refold the sequences. \textbf{Diversity}, measured as the average of one minus the pairwise TM-Score~\citep{zhang2004scoring}, indicates structural dissimilarity among designed peptides.

\paragraph{Results.} Tab.~\ref{tab:co-design} illustrates that our SurfFlow generates significantly more diversified and consistent peptides with better binding energy and closer resemblance compared to the baselines. Specifically, SurfFlow achieves the state-of-the-art AAR of 54.07\% and RMSD of 1.96~\AA with improvements of 5.51\% and 5.31\% over the full-atom PepFlow. Besides, it also owns a stronger capacity for designing peptides with a more accurate binding ratio of 87.38\% and a higher affinity proportion of 22.51\% (see Fig.~\ref{fig:energy}). These statistics confirm the benefits of explicitly modeling the surface geometry and biochemical constraints. It is worth mentioning that the decoupled approach, RFDiffusion, attains better Stability (26.82\% v.s. 22.46\%) and Designability (78.52\% v.s. 73.60\%), as it is trained in the entire PDB and is prone towards structures with more stable motifs.   

\begin{figure}[t]
    \centering
    \includegraphics[width=1.0\linewidth]{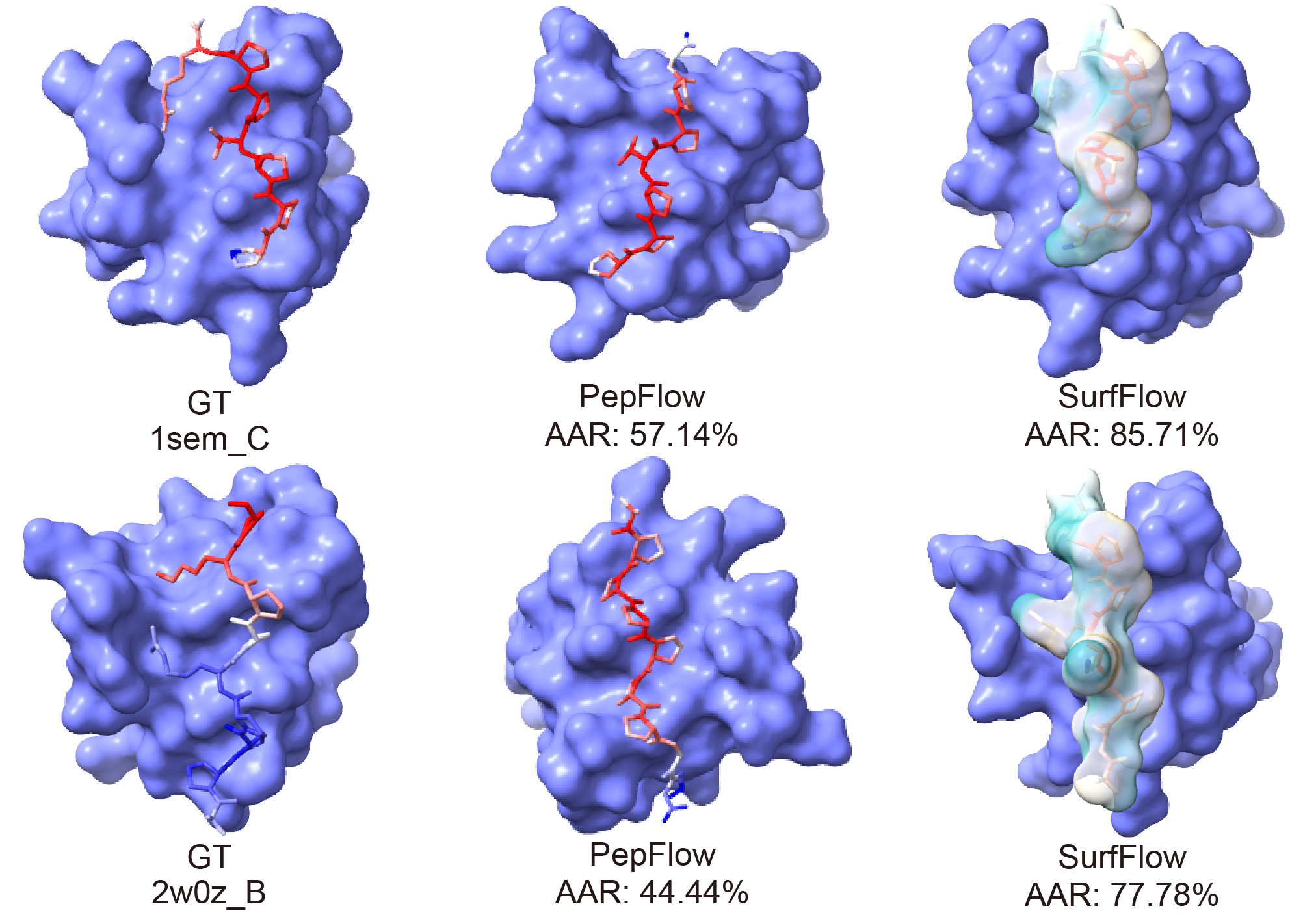}
    \vspace{-0.5em}
    \caption{Peptide designed by DL algorithms and references.}
    \Description{ }
    \vspace{-0.5em}
    \label{fig: uncondition_case}
\end{figure}

Fig.~\ref{fig: uncondition_case} presents two examples of designed peptides by full-atom PepFlow and surface-based SurfFlow and App.~\ref{app:visual_evo} shows how the positions of surface point clouds evolve over the entire period. SurfFlow generates peptides with topologically similar geometries, irrespective of their native length. Notably, produced peptides share comparable side-chain compositions and conformations, enabling efficient interactions with the target protein at the appropriate binding site.

We also investigate the contributions of SurfFlow's components in Tab.~\ref{tab:co-design}, where a naive PointNet~\citep{qi2017pointnet} is used as the replacement of ESGN. It shows that ESGN's effective modeling of surface interactions is the primary key to SurfFlow's success.
The removal of the biophysical features significantly reduces the performance with a drop of 3.76\% in Designability and 4.91\% in Diversity. Besides, it also indicates that the inclusion of surface orientation is beneficial, which brings an improvement of 1.94\% in AAR. We also report the surface consistency in Tab.~\ref{tab:surf} and inference computational costs in App.~\ref{tab:time}, and SurfFlow substantially improves the surface similarity between generated peptides and ground truth with comparable complexity. 
\begin{figure}[t]
    \centering
    \includegraphics[width=0.7\linewidth]{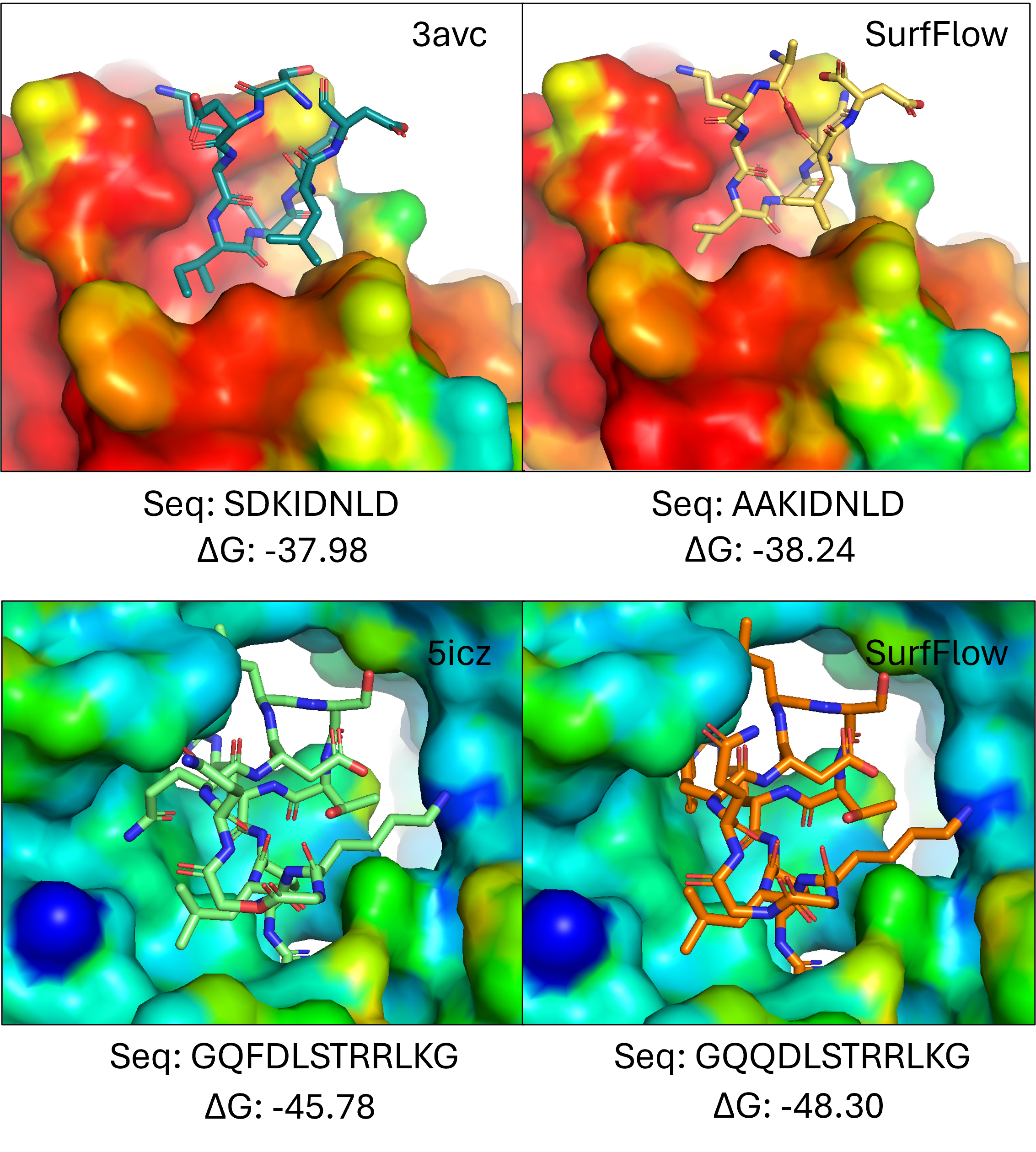}
    \vspace{-0.5em}
    \caption{Peptide design with the cyclic condition.}
    \Description{ }
    \label{fig: condition_case}
    \vspace{-0.5em}
\end{figure}

\subsection{Conditioned Sequence-structure Co-design}
There are several key characteristics of effective peptides widely recognized in the biological community. Cyclicity, for instance, is are polypeptide chains with a circular bond sequence. Many occur naturally and exhibit antimicrobial or toxic properties, while others are laboratory-synthesized~\citep{jensen2009peptide}. These cyclic peptides typically demonstrate high resistance to digestion, making them attractive to researchers developing novel oral medications like antibiotics and immunosuppressants~\citep{craik2006seamless}. Additionally, evidence indicates that disulfide patterns are crucial in the folding and structural stabilization of peptides. The deliberate introduction of disulfide bridges into natural or engineered peptides can often enhance their biological activities, specificities, and stabilities~\citep{annis199710}. Given these insights, we propose a conditional co-design challenge. This challenge involves calculating the proportions of cyclic peptides and peptides containing disulfide bonds within generated peptide sets for evaluation purposes.

\paragraph{Results.} Tab.~\ref{tab:condition} documents the results. Without conditional constraints, neither full-atom deep generative models nor our omni-design method met the requirements for cyclicity or disulfide bridges. Even with the incorporation of molecular surfaces, only 2-4\% of the 189 designed peptides in the test samples exhibited cyclicity or contained disulfide bonds. In contrast, when trained with additional conditions and prompted to generate peptides with specific properties, SurfFlow significantly increases the proportion of peptides with desired characteristics, with cyclicity from 2.67\% to 8.02\% and disulfide bonds from 4.27\% to 9.10\%. Fig.~\ref{fig: condition_case} visualizes two examples with the cyclic condition. Interestingly, in the case of 3AVC, where the native peptide is non-cyclic, SurfFlow generated a novel cyclic peptide with a lower binding energy ($\Delta G = -37.98$ v.s. $\Delta G = -38.24$). For 5ICZ, which originally had a cyclic peptide, SurfFlow designed a new type of cyclic peptide with a better binding energy ($\Delta G = -45.78$ v.s. $\Delta G = -48.30$). 
\begin{table}[t]
\centering
\caption{Evaluation on the surface consistency on PepMerge. }
\label{tab:surf} 
\resizebox{0.6\columnwidth}{!}{
\begin{tabular}{l|ccc} \toprule
    \textbf{Models} & \textbf{IoU} ($\uparrow$) & \textbf{CD} ($\downarrow$) & \textbf{NC}($\uparrow$) \\ \midrule
    PepFlow &  0.7252 & 9.437 & 0.3088 \\
    \grow{SurfFlow} & \textbf{0.8104} & \textbf{6.347} & \textbf{0.3859}\\ \bottomrule
\end{tabular}}
\vspace{-1em}
\end{table}
\begin{table}[t]
\caption{Proportions of cyclic peptides and peptides with disulfide bonds designed by different mechanisms and in the original PepMerge dataset.}
\centering
\resizebox{1.0\columnwidth}{!}{
\begin{tabular}{l|ccc|c} \toprule
    \textbf{Metrics} & PepFlow & SurfFlow (w/o $c$) & \cellcolor{blue!25}SurfFlow & PepMerge \\ \midrule
    \textbf{Cyclicity}& 2.13\% & 2.67\% & \cellcolor{blue!25}\textbf{8.02\%} &  15.50\% \\ 
    \textbf{Disulfide Bond}\% & 3.21\% & 4.27\%  & \cellcolor{blue!25}\textbf{9.10\%} & 18.18\% \\ \bottomrule
\end{tabular}}
\vspace{-1em}
\label{tab:condition} 
\end{table}

\subsection{Side-chain Packing}
Side-chain packing is a critical task in protein structure modeling, focusing on predicting peptide side-chain angles. Our approach generates 64 distinct side-chain conformations for each peptide using multiple models, employing a partial sampling strategy to efficiently recover the most probable side-chain angles. This method allows us to navigate the conformational space effectively while reducing computational overhead and maintaining high accuracy.

\paragraph{Baselines.} We compare SurfFlow against several established approaches. Energy-based methods such as RosettaPacker~\citep{leman2020macromolecular} and SCWRL4~\citep{krivov2009improved} rely on physical energy minimization techniques for side-chain positioning. Additionally, learning-based models, including DLPacker~\citep{misiura2022dlpacker}, AttnPacker~\citep{mcpartlon2023end}, and DiffPack~\citep{zhang2024diffpack} leverage various DL strategies like attention mechanisms and diffusion to predict side-chain configurations based on learned representations of protein structure.

\paragraph{Metrics.} Two metrics are employed for evaluation. First, we calculate the Mean Absolute Error (MAE) of 4 key torsion angles: $\{\boldsymbol{\chi}_i\}_{i=1}^4$. Given the inherent flexibility of side chains and the importance of small deviations in structural biology, we also report the proportion of predictions that fall within a $20^{\circ}$ deviation from the ground truth. This additional metric captures the practical accuracy of side-chain prediction, emphasizing how well models perform within biologically relevant error margins. 

\paragraph{Results.} Tab.~\ref{tab: side-chain} reports the results; it can be observed that the incorporation of molecular surfaces contributes to more accurate predictions of all four side-chain angles compared to full-atom models and other baselines. SurfFlow attains the highest correct ratio of 63.02\%. The enhanced accuracy suggests that surface information provides crucial context for predicting side-chain configurations, likely by better representing the local environment and potential interactions that influence side-chain positioning.

\begin{table}[t]
\caption{Evaluation results of the side-chain packing task.}
\vspace{-0.5em}
\centering
\resizebox{0.85\columnwidth}{!}{%
\begin{tabular}{lccccc}\toprule
 & \multicolumn{4}{c}{ MSE $^{\circ} \downarrow$} & \\
\cline { 2 - 5 } & $\chi_1$ & $\chi_2$ & $\chi_3$ & $\chi_4$ & Correct $\% \uparrow$ \\ \midrule
Rosseta & 38.31 & 43.23 & 53.61 & 71.67 & 57.03 \\
SCWRL4 & 30.06 & 40.40 & 49.71 & 53.79 & 60.54 \\
DLPacker & 22.44 & 35.65 & 58.53 & 61.70 & 60.91 \\
AttnPacker & 19.04 & 28.49 & 40.16 & 60.04 & 61.46 \\
DiffPack & 17.92 & 26.08 & 36.20 & 67.82 & 62.58 \\
PepFlow & \underline{17.38} & \underline{24.71} & \underline{33.63} & \underline{58.49} & \underline{62.79} \\ \midrule
\grow{SurfFlow} & \textbf{17.13}  & \textbf{23.86} & \textbf{31.97} & \textbf{55.08} & \textbf{63.02} \\  \bottomrule
\end{tabular}}
\vspace{-1em}
\label{tab: side-chain}
\end{table}

\section{Conclusion}
This work presents SurfFlow, a novel model that produces all protein modalities -- sequence, structure, and surface -- concurrently. We apply SurfFlow to solve a specific peptide design challenge and integrate key characteristics like cyclicity and disulfide bonds into the generation process. Empirical results prove the reasonability and promise of considering molecular surfaces for protein discovery. Limitations and future work are elucidated in App.~\ref{app:feature_work}. 

\section*{Acknowledgement}
This work is supported by the Beijing Municipal Science \& Technology Commission and the Zhongguancun Administrative Committee under project number Z221100003522019.

\bibliographystyle{ACM-Reference-Format}
\bibliography{cite}

@article{gasteiger2020fast,
  title={Fast and uncertainty-aware directional message passing for non-equilibrium molecules},
  author={Gasteiger, Johannes and Giri, Shankari and Margraf, Johannes T and G{\"u}nnemann, Stephan},
  journal={arXiv preprint arXiv:2011.14115},
  year={2020}
}

@article{zhang2023equipocket,
  title={Equipocket: an e (3)-equivariant geometric graph neural network for ligand binding site prediction},
  author={Zhang, Yang and others},
  journal={arXiv preprint arXiv:2302.12177},
  year={2023}
}

@article{lai2024interformer,
  title={Interformer: An Interaction-Aware Model for Protein-Ligand Docking and Affinity Prediction},
  author={Lai, Houtim and others},
  year={2024}
}

@article{morrison2006lock,
  title={A lock-and-key model for protein--protein interactions},
  author={Morrison, Julie L and others},
  journal={Bioinformatics},
  volume={22},
  number={16},
  pages={2012--2019},
  year={2006},
  publisher={Oxford University Press}
}

@article{jumper2021highly,
  title={Highly accurate protein structure prediction with AlphaFold},
  author={Jumper, John and others},
  journal={Nature},
  volume={596},
  number={7873},
  pages={583--589},
  year={2021},
  publisher={Nature Publishing Group}
}

@article{wu2022pre,
  title={Pre-Training of Equivariant Graph Matching Networks with Conformation Flexibility for Drug Binding},
  author={Wu, Fang and others},
  journal={Advanced Science},
  volume={9},
  number={33},
  pages={2203796},
  year={2022},
  publisher={Wiley Online Library}
}

@article{li2024full,
  title={Full-Atom Peptide Design based on Multi-modal Flow Matching},
  author={Li, Jiahan and others},
  journal={arXiv preprint arXiv:2406.00735},
  year={2024}
}

@article{song2024surfpro,
  title={SurfPro: Functional Protein Design Based on Continuous Surface},
  author={Song, Zhenqiao and Huang, Tinglin and Li, Lei and Jin, Wengong},
  journal={arXiv preprint arXiv:2405.06693},
  year={2024}
}

@inproceedings{qi2017pointnet,
  title={Pointnet: Deep learning on point sets for 3d classification and segmentation},
  author={Qi, Charles R and Su, Hao and Mo, Kaichun and Guibas, Leonidas J},
  booktitle={Proceedings of the IEEE conference on computer vision and pattern recognition},
  pages={652--660},
  year={2017}
}

@article{tang2024bc,
  title={BC-Design: A Biochemistry-Aware Framework for Highly Accurate Inverse Protein Folding},
  author={Tang, Xiangru and others},
  journal={bioRxiv},
  pages={2024--10},
  year={2024},
  publisher={Cold Spring Harbor Laboratory}
}

@article{wu2024d,
  title={D-Flow: Multi-modality Flow Matching for D-peptide Design},
  author={Wu, Fang and others},
  journal={arXiv preprint arXiv:2411.10618},
  year={2024}
}

@article{gainza2020deciphering,
  title={Deciphering interaction fingerprints from protein molecular surfaces using geometric deep learning},
  author={Gainza, Pablo and others},
  journal={Nature Methods},
  volume={17},
  number={2},
  pages={184--192},
  year={2020},
  publisher={Nature Publishing Group US New York}
}

@article{somnath2021multi,
  title={Multi-scale representation learning on proteins},
  author={Somnath, Vignesh Ram and Bunne, Charlotte and Krause, Andreas},
  journal={Advances in Neural Information Processing Systems},
  volume={34},
  pages={25244--25255},
  year={2021}
}

@article{lisanza2023joint,
  title={Joint generation of protein sequence and structure with RoseTTAFold sequence space diffusion},
  author={Lisanza, Sidney Lyayuga and others},
  journal={bioRxiv},
  pages={2023--05},
  year={2023},
  publisher={Cold Spring Harbor Laboratory}
}

@inproceedings{zhang2023adding,
  title={Adding conditional control to text-to-image diffusion models},
  author={Zhang, Lvmin and Rao, Anyi and Agrawala, Maneesh},
  booktitle={Proceedings of the IEEE/CVF International Conference on Computer Vision},
  pages={3836--3847},
  year={2023}
}

@article{watson2023novo,
  title={De novo design of protein structure and function with RFdiffusion},
  author={Watson, Joseph L and others},
  journal={Nature},
  volume={620},
  number={7976},
  pages={1089--1100},
  year={2023},
  publisher={Nature Publishing Group UK London}
}

@book{norris1998markov,
  title={Markov chains},
  author={Norris, James R},
  number={2},
  year={1998},
  publisher={Cambridge university press}
}

@article{yim2023fast,
  title={Fast protein backbone generation with SE (3) flow matching},
  author={Yim, Jason and  others},
  journal={arXiv preprint arXiv:2310.05297},
  year={2023}
}

@article{zheng2023guided,
  title={Guided flows for generative modeling and decision making},
  author={Zheng, Qinqing and others},
  journal={arXiv preprint arXiv:2311.13443},
  year={2023}
}

@article{ramesh2022hierarchical,
  title={Hierarchical text-conditional image generation with clip latents},
  author={Ramesh, Aditya and Dhariwal, Prafulla and Nichol, Alex and Chu, Casey and Chen, Mark},
  journal={arXiv preprint arXiv:2204.06125},
  volume={1},
  number={2},
  pages={3},
  year={2022}
}

@article{dhariwal2021diffusion,
  title={Diffusion models beat gans on image synthesis},
  author={Dhariwal, Prafulla and Nichol, Alexander},
  journal={Advances in neural information processing systems},
  volume={34},
  pages={8780--8794},
  year={2021}
}

@article{annis199710,
  title={Disulfide bond formation in peptides},
  author={Annis, Ioana and Hargittai, Balazs and Barany, George},
  journal={Methods in enzymology},
  volume={289},
  pages={198--221},
  year={1997},
  publisher={Elsevier}
}

@article{buckton2021cyclic,
  title={Cyclic peptides as drugs for intracellular targets: the next frontier in peptide therapeutic development},
  author={Buckton, Laura K and Rahimi, Marwa N and McAlpine, Shelli R},
  journal={Chemistry--A European Journal},
  volume={27},
  number={5},
  pages={1487--1513},
  year={2021},
  publisher={Wiley Online Library}
}

@article{jones1996principles,
  title={Principles of protein-protein interactions.},
  author={Jones, Susan and Thornton, Janet M},
  journal={Proceedings of the National Academy of Sciences},
  volume={93},
  number={1},
  pages={13--20},
  year={1996},
  publisher={National Acad Sciences}
}

@article{kastritis2013binding,
  title={On the binding affinity of macromolecular interactions: daring to ask why proteins interact},
  author={Kastritis, Panagiotis L and Bonvin, Alexandre MJJ},
  journal={Journal of The Royal Society Interface},
  volume={10},
  number={79},
  pages={20120835},
  year={2013},
  publisher={The Royal Society}
}

@article{ramasubramanian2024hybrid,
  title={A Hybrid Diffusion Model for Stable, Affinity-Driven, Receptor-Aware Peptide Generation},
  author={Ramasubramanian, Vishva Saravanan and Choudhuri, Soham and Ghosh, Bhaswar},
  journal={bioRxiv},
  pages={2024--03},
  year={2024},
  publisher={Cold Spring Harbor Laboratory}
}

@article{vanhee2011computational,
  title={Computational design of peptide ligands},
  author={Vanhee, Peter and van der Sloot, Almer M and Verschueren, Erik and Serrano, Luis and Rousseau, Frederic and Schymkowitz, Joost},
  journal={Trends in biotechnology},
  volume={29},
  number={5},
  pages={231--239},
  year={2011},
  publisher={Elsevier}
}

@article{ho2020denoising,
  title={Denoising diffusion probabilistic models},
  author={Ho, Jonathan and Jain, Ajay and Abbeel, Pieter},
  journal={Advances in neural information processing systems},
  volume={33},
  pages={6840--6851},
  year={2020}
}

@article{guan20233d,
  title={3d equivariant diffusion for target-aware molecule generation and affinity prediction},
  author={Guan, Jiaqi and Qian, Wesley Wei and Peng, Xingang and Su, Yufeng and Peng, Jian and Ma, Jianzhu},
  journal={arXiv preprint arXiv:2303.03543},
  year={2023}
}

@article{bhardwaj2016accurate,
  title={Accurate de novo design of hyperstable constrained peptides},
  author={Bhardwaj, Gaurav and Mulligan, Vikram Khipple and Bahl, Christopher D and Gilmore, Jason M and Harvey, Peta J and Cheneval, Olivier and Buchko, Garry W and Pulavarti, Surya VSRK and Kaas, Quentin and Eletsky, Alexander and others},
  journal={Nature},
  volume={538},
  number={7625},
  pages={329--335},
  year={2016},
  publisher={Nature Publishing Group UK London}
}

@article{muttenthaler2021trends,
  title={Trends in peptide drug discovery},
  author={Muttenthaler, Markus and King, Glenn F and Adams, David J and Alewood, Paul F},
  journal={Nature reviews Drug discovery},
  volume={20},
  number={4},
  pages={309--325},
  year={2021},
  publisher={Nature Publishing Group UK London}
}

@article{wang2022therapeutic,
  title={Therapeutic peptides: current applications and future directions},
  author={Wang, Lei and others},
  journal={Signal transduction and targeted therapy},
  volume={7},
  number={1},
  pages={48},
  year={2022},
  publisher={Nature Publishing Group UK London}
}

@article{wu2024hierarchical,
  title={A hierarchical training paradigm for antibody structure-sequence co-design},
  author={Wu, Fang and Li, Stan Z},
  journal={Advances in Neural Information Processing Systems},
  volume={36},
  year={2024}
}

@article{martinkus2024abdiffuser,
  title={AbDiffuser: full-atom generation of in-vitro functioning antibodies},
  author={Martinkus, Karolis and others},
  journal={Advances in Neural Information Processing Systems},
  volume={36},
  year={2024}
}

@article{misiura2022dlpacker,
  title={DLPacker: Deep learning for prediction of amino acid side chain conformations in proteins},
  author={Misiura, Mikita and others},
  journal={Proteins: Structure, Function, and Bioinformatics},
  volume={90},
  number={6},
  pages={1278--1290},
  year={2022},
  publisher={Wiley Online Library}
}

@article{krivov2009improved,
  title={Improved prediction of protein side-chain conformations with SCWRL4},
  author={Krivov, Georgii G and Shapovalov, Maxim V and Dunbrack Jr, Roland L},
  journal={Proteins: Structure, Function, and Bioinformatics},
  volume={77},
  number={4},
  pages={778--795},
  year={2009},
  publisher={Wiley Online Library}
}

@article{leman2020macromolecular,
  title={Macromolecular modeling and design in Rosetta: recent methods and frameworks},
  author={Leman, Julia Koehler and others},
  journal={Nature methods},
  volume={17},
  number={7},
  pages={665--680},
  year={2020},
  publisher={Nature Publishing Group US New York}
}

@article{zhang2024diffpack,
  title={Diffpack: A torsional diffusion model for autoregressive protein side-chain packing},
  author={Zhang, Yangtian and others},
  journal={Advances in Neural Information Processing Systems},
  volume={36},
  year={2024}
}

@article{mcpartlon2023end,
  title={An end-to-end deep learning method for protein side-chain packing and inverse folding},
  author={McPartlon, Matthew and Xu, Jinbo},
  journal={Proceedings of the National Academy of Sciences},
  volume={120},
  number={23},
  pages={e2216438120},
  year={2023},
  publisher={National Acad Sciences}
}

@article{laskowski1996protein,
  title={Protein clefts in molecular recognition and function.},
  author={Laskowski, Roman A and others},
  journal={Protein science: a publication of the Protein Society},
  volume={5},
  number={12},
  pages={2438},
  year={1996},
  publisher={Wiley}
}

@article{bose2023se,
  title={Se (3)-stochastic flow matching for protein backbone generation},
  author={Bose, Avishek Joey and others},
  journal={arXiv preprint arXiv:2310.02391},
  year={2023}
}

@inproceedings{lee2023pre,
  title={Pre-training Sequence, Structure, and Surface Features for Comprehensive Protein Representation Learning},
  author={Lee, Youhan and others},
  booktitle={The Twelfth International Conference on Learning Representations},
  year={2023}
}

@article{gainza2023novo,
  title={De novo design of protein interactions with learned surface fingerprints},
  author={Gainza, Pablo and others},
  journal={Nature},
  volume={617},
  number={7959},
  pages={176--184},
  year={2023},
  publisher={Nature Publishing Group UK London}
}

@article{albergo2022building,
  title={Building normalizing flows with stochastic interpolants},
  author={Albergo, Michael S and Vanden-Eijnden, Eric},
  journal={arXiv preprint arXiv:2209.15571},
  year={2022}
}

@article{albergo2023stochastic,
  title={Stochastic interpolants: A unifying framework for flows and diffusions},
  author={Albergo, Michael S and Boffi, Nicholas M and Vanden-Eijnden, Eric},
  journal={arXiv preprint arXiv:2303.08797},
  year={2023}
}

@article{blanco2021tutorial,
  title={A tutorial on SE(3) transformation parameterizations and on-manifold optimization},
  author={Blanco-Claraco, Jos{\'e} Luis},
  journal={arXiv preprint arXiv:2103.15980},
  year={2021}
}

@book{lee2018introduction,
  title={Introduction to Riemannian manifolds},
  author={Lee, John M},
  volume={2},
  year={2018},
  publisher={Springer}
}

@article{sun2024dsr,
  title={DSR: dynamical surface representation as implicit neural networks for protein},
  author={Sun, Daiwen and Huang, He and Li, Yao and Gong, Xinqi and Ye, Qiwei},
  journal={Advances in Neural Information Processing Systems},
  volume={36},
  year={2024}
}

@inproceedings{sverrisson2021fast,
  title={Fast end-to-end learning on protein surfaces},
  author={Sverrisson, Freyr and Feydy, Jean and Correia, Bruno E and Bronstein, Michael M},
  booktitle={Proceedings of the IEEE/CVF Conference on Computer Vision and Pattern Recognition},
  pages={15272--15281},
  year={2021}
}

@inproceedings{wu2024surface,
  title={Surface-VQMAE: Vector-quantized Masked Auto-encoders on Molecular Surfaces},
  author={Wu, Fang and Li, Stan Z},
  booktitle={International Conference on Machine Learning},
  pages={53619--53634},
  year={2024},
  organization={PMLR}
}

@article{wu2024protein,
  title={Protein structure generation via folding diffusion},
  author={Wu, Kevin E and others},
  journal={Nature communications},
  volume={15},
  number={1},
  pages={1059},
  year={2024},
  publisher={Nature Publishing Group UK London}
}

@article{dauparas2022robust,
  title={Robust deep learning--based protein sequence design using ProteinMPNN},
  author={Dauparas and others},
  journal={Science},
  volume={378},
  number={6615},
  pages={49--56},
  year={2022},
  publisher={American Association for the Advancement of Science}
}

@article{kong2024full,
  title={Full-atom peptide design with geometric latent diffusion},
  author={Kong, Xiangzhe and Huang, Wenbing and Liu, Yang},
  journal={arXiv preprint arXiv:2402.13555},
  year={2024}
}

@article{robinson2014msm,
  title={MSM: a new flexible framework for multimodal surface matching},
  author={Robinson and others},
  journal={Neuroimage},
  volume={100},
  pages={414--426},
  year={2014},
  publisher={Elsevier}
}

@article{delano2002pymol,
  title={Pymol: An open-source molecular graphics tool},
  author={DeLano, Warren L and others},
  journal={CCP4 Newsl. Protein Crystallogr},
  volume={40},
  number={1},
  pages={82--92},
  year={2002},
  publisher={Citeseer}
}

@article{liu2022flow,
  title={Flow straight and fast: Learning to generate and transfer data with rectified flow},
  author={Liu, Xingchao and Gong, Chengyue and Liu, Qiang},
  journal={arXiv preprint arXiv:2209.03003},
  year={2022}
}

@article{lipman2022flow,
  title={Flow matching for generative modeling},
  author={Lipman, Yaron and others},
  journal={arXiv preprint arXiv:2210.02747},
  year={2022}
}

@article{fisher2001lehninger,
  title={Lehninger principles of biochemistry, ; by David L. Nelson and Michael M. Cox},
  author={Fisher, Matthew},
  journal={The Chemical Educator},
  volume={6},
  pages={69--70},
  year={2001},
  publisher={Springer}
}

@article{jensen2009peptide,
  title={Peptide and Protein Design for Biopharmaceutical Applications},
  author={Jensen, Knud J},
  year={2009},
  publisher={Wiley Online Library}
}

@article{craik2006seamless,
  title={Seamless proteins tie up their loose ends},
  author={Craik, David J},
  journal={Science},
  volume={311},
  number={5767},
  pages={1563--1564},
  year={2006},
  publisher={American Association for the Advancement of Science}
}

@article{lin2024ppflow,
  title={PPFLOW: Target-Aware Peptide Design with Torsional Flow Matching},
  author={Lin, Haitao and others},
  journal={bioRxiv},
  pages={2024--03},
  year={2024},
  publisher={Cold Spring Harbor Laboratory}
}

@article{gat2024discrete,
  title={Discrete Flow Matching},
  author={Gat, Itai and others},
  journal={arXiv preprint arXiv:2407.15595},
  year={2024}
}

@article{campbell2024generative,
  title={Generative flows on discrete state-spaces: Enabling multimodal flows with applications to protein co-design},
  author={Campbell, Andrew and others},
  journal={arXiv preprint arXiv:2402.04997},
  year={2024}
}

@article{alford2017rosetta,
  title={The Rosetta all-atom energy function for macromolecular modeling and design},
  author={Alford, Rebecca F and others},
  journal={Journal of chemical theory and computation},
  volume={13},
  number={6},
  pages={3031--3048},
  year={2017},
  publisher={ACS Publications}
}

@article{wen2019pepbdb,
  title={PepBDB: a comprehensive structural database of biological peptide--protein interactions},
  author={Wen, Zeyu and He, Jiahua and Tao, Huanyu and Huang, Sheng-You},
  journal={Bioinformatics},
  volume={35},
  number={1},
  pages={175--177},
  year={2019},
  publisher={Oxford University Press}
}

@article{steinegger2017mmseqs2,
  title={MMseqs2 enables sensitive protein sequence searching for the analysis of massive data sets},
  author={Steinegger, Martin and S{\"o}ding, Johannes},
  journal={Nature biotechnology},
  volume={35},
  number={11},
  pages={1026--1028},
  year={2017},
  publisher={Nature Publishing Group US New York}
}

@article{wei2024q,
  title={Q-biolip: A comprehensive resource for quaternary structure-based protein--ligand interactions},
  author={Wei, Hong and Wang, Wenkai and Peng, Zhenling and Yang, Jianyi},
  journal={Genomics, Proteomics \& Bioinformatics},
  volume={22},
  number={1},
  year={2024},
  publisher={Oxford Academic}
}

@article{sun2022score,
  title={Score-based continuous-time discrete diffusion models},
  author={Sun, Haoran and others},
  journal={arXiv preprint arXiv:2211.16750},
  year={2022}
}

@article{stark2024dirichlet,
  title={Dirichlet flow matching with applications to dna sequence design},
  author={Stark, Hannes and others},
  journal={arXiv preprint arXiv:2402.05841},
  year={2024}
}

@article{luo2022antigen,
  title={Antigen-specific antibody design and optimization with diffusion-based generative models},
  author={Luo, Shitong and Su, Yufeng and Peng, Xingang and Wang, Sheng and Peng, Jian and Ma, Jianzhu},
  journal={bioRxiv},
  year={2022},
  publisher={Cold Spring Harbor Laboratory}
}

@article{lin2022language,
  title={Language models of protein sequences at the scale of evolution enable accurate structure prediction},
  author={Lin, Zeming and Akin, Halil and Rao, Roshan and Hie, Brian and Zhu, Zhongkai and Lu, Wenting and dos Santos Costa, Allan and Fazel-Zarandi, Maryam and Sercu, Tom and Candido, Sal and others},
  journal={bioRxiv},
  year={2022},
  publisher={Cold Spring Harbor Laboratory}
}

@inproceedings{satorras2021n,
  title={E (n) equivariant graph neural networks},
  author={Satorras, V{\i}ctor Garcia and Hoogeboom, Emiel and Welling, Max},
  booktitle={International conference on machine learning},
  pages={9323--9332},
  year={2021},
  organization={PMLR}
}

@article{ganea2021independent,
  title={Independent se (3)-equivariant models for end-to-end rigid protein docking},
  author={Ganea, Octavian-Eugen and others},
  journal={arXiv preprint arXiv:2111.07786},
  year={2021}
}

@article{zhang2004scoring,
  title={Scoring function for automated assessment of protein structure template quality},
  author={Zhang, Yang and Skolnick, Jeffrey},
  journal={Proteins: Structure, Function, and Bioinformatics},
  volume={57},
  number={4},
  pages={702--710},
  year={2004},
  publisher={Wiley Online Library}
}

\appendix
\onecolumn

\section{Parameterization with Networks}
\label{app:networks}
In order to model the joint distribution of the peptide based on its target protein $p\left(C^{\mathrm{pep}} \mid C^{\mathrm{rec}}\right)$, we adopt an encoder-decoder framework to generate peptides. The encoder extracts the geometric and biochemical features of the receptor $C^{\mathrm{rec}}$ as a condition of the generation process, while the decoder regresses the vector fields of our multi-modality flow matching architecture. 

\paragraph{Encoder.} We first utilize a time-independent equivariant geometric encoder to capture the context information of the receptor. Specifically, it takes the sequence and structure of the target protein $C_{\mathrm{rec}}$ and computes the hidden residue representations $\boldsymbol{h}_{\mathrm{rec}}\in\mathbb{R}^{n_{\mathrm{rec}}\times \psi_{\mathrm{rec}}}$ and the residue-pair embeddings $\boldsymbol{z}_{\mathrm{rec}}\in\mathbb{R}^{n_{\mathrm{rec}}\times \psi_{\mathrm{pair}}}$.

\paragraph{Decoder.} The decoder receives $(\boldsymbol{h}_{\mathrm{rec}}, \boldsymbol{z}_{\mathrm{rec}})$ and is time-dependent. It consists of two geometric networks: one is the 6-layer Invariant Point Attention (IPA) module~\citep{jumper2021highly} to regress the vector fields of the internal structures $\left\{\left(a_j, O_j, \boldsymbol{x}_j,\boldsymbol{\chi}_j\right)\right\}_{j=1}^{n_{\mathrm{pep}}}$, and the other is an variant of equivariant graph neural networks (EGNN)~\citep{satorras2021n} to regress the vector fields of the surface geometry $\left\{\left(\boldsymbol{x}_i^s, \boldsymbol{n}_i^s, \boldsymbol{\tau}_i^s, \boldsymbol{\Upsilon}_i^s\right)\right\}_{i=1}^m$. To be specific, we first sample a random timestep $t\sim \mathcal{U}(0,1)$ to construct the time-dependent vector fields for every modality of the peptide $C_{\mathrm{pep}}$, containing sequence, structure, and surface. Both IPA and EGNN take the timestamp $t$, the interplant state of peptide's internal structure $\left({a}_t, {O}_t, \boldsymbol{x}_t, \boldsymbol{\chi}_t\right)$ as well as receptor's information $(\boldsymbol{h}_{\mathrm{rec}}, \boldsymbol{z}_{\mathrm{rec}})$ as input, and the interplant state of peptide's surface $\left(\boldsymbol{x}_t^s, \boldsymbol{n}_t^s, \boldsymbol{\tau}_t^s, \boldsymbol{\Upsilon}_t^s\right)$ is also forwarded into EGNN.  Subsequently, IPA recovers the internal structure of original peptide $\left(\hat{{a}}_1, \hat{{O}}_1, \hat{\boldsymbol{x}}_1, \hat{\boldsymbol{\chi}}_1\right)$, while the 3-layer EGNN recovers the surface $\left(\hat{\boldsymbol{x}}_1^s, \hat{\boldsymbol{n}}_1^s, \hat{\boldsymbol{\tau}}_1^s, \hat{\boldsymbol{\Upsilon}}_1^s\right)$. Moreover, two additional losses containing the backbone position loss $\mathcal{L}_{\mathrm{bb}}(\theta)$ and the torsion angle loss $\mathcal{L}_{\mathrm{tor}}(\theta)$ is imposed for extra constraint. 

\paragraph{Equivariance.} The joint distribution $p\left(C^{\mathrm{pep}} \mid C^{\mathrm{rec}}\right)$ must satisfy the roto-translational equivariance to ensure the generalization. That is, for any translation vector $\boldsymbol{\epsilon} \in \mathbb{R}^3$ and for any orthogonal matrix $O \in \mathbb{R}^{3 \times 3}$, it should satisfy:
\begin{equation}
    p\left(OC^{\mathrm{pep}} + \boldsymbol{\epsilon}\mid OC^{\mathrm{rec}} + \boldsymbol{\epsilon}\right),
\end{equation}
where $OC^{\mathrm{pep}} + \boldsymbol{\epsilon} = \left\{\left(a_j, O_j, O\boldsymbol{x}_j + \boldsymbol{\epsilon}, \boldsymbol{\chi}_j\right)\right\}_{j=1}^{n_{\mathrm{pep}}} \cup \left\{\left(O\boldsymbol{x}_i^s + \boldsymbol{\epsilon}, \boldsymbol{n}_i^s, \boldsymbol{\tau}_i^s, \boldsymbol{\Upsilon}_i^s\right)\right\}_{i=1}^m$. Using zero-mass-center, we subtract the mass center of the receptor from all inputs' coordinates to achieve the invariance to translation, which also improves the training stability. Moreover, it can be proven that when the prior distributions $p(a_0)$, $p(O_0)$, $p(\boldsymbol{x}_0)$, $p(\boldsymbol{\chi}_0)$, $p(\boldsymbol{x}_0^s)$, $p(\boldsymbol{n}_0^s)$, $p(\boldsymbol{\tau}_0^s)$, and $p(\boldsymbol{\Upsilon}_0^s)$ are $\mathrm{SE}(3)$-invariant, while the vector fields $v^{\text{cat}}(\cdot)$ and $v^{\text{con}}(\cdot)$ are $\mathrm{SE}(3)$-invariant, and the vector field $v^{\text{ori}}(\cdot)$ is $\mathrm{SO}(3)$-equivariant and $\mathrm{T}(3)$-invariant, and the vector field $v^{\text{pos}}(\cdot)$ is $\mathrm{SE}(3)$-equivariant, then the density $p\left(C^{\mathrm{pep}} \mid C^{\mathrm{rec}}\right)$ generated by the ODE sampling process is $\mathrm{SE}(3)$-equivariant. Notably, the choice of IPA and EGNN guarantees the equivariance and invariance requirement of those vector fields. 

\paragraph{Conditional Design.} Controlling the output of deep generative models such as diffusions or flows has become a hotspot in recent years. Apart from the receptor $C_{\mathrm{rec}}$, we often want to create peptides with specific conditions, such as cyclicity and disulfide bonds. Conventionally, conditional controls involve classifier guidance~\citep{dhariwal2021diffusion} and classifier-free guidance~\citep{ho2020denoising}. Evidence indicates that with enough high-quality training data, classifier-free guidance tends to yield better results, being able to generate an almost infinite number of sample categories without the need to retrain a classifier architecture. This leads to wide usage of classifier-free guidance in modern AI systems~\citep{ramesh2022hierarchical,zheng2023guided}. Due to this superiority, we adopt the classifier-free guidance to realize the insertion of peptide conditions $c$ and transfer $p(C_{\mathrm{pep}}|C_{\mathrm{rec}})$ to $p(C_{\mathrm{pep}}|C_{\mathrm{rec}}, c)$. Specifically, we denote the null condition by $\emptyset$ by convention and set $p(C_{\mathrm{pep}}|C_{\mathrm{rec}}, \emptyset) := p(C_{\mathrm{pep}}|C_{\mathrm{rec}})$ and $u_t(\cdot | \cdot, \emptyset) :=u_t(\cdot | \cdot)$. Then taking the surface point positions for example, our training loss becomes~\citep{zheng2023guided}:
\begin{equation}
    \mathcal{L}_{\mathrm{pos}}(\theta)=\mathbb{E}_{t\sim \mathcal{U}(0,1), b, p\left(\boldsymbol{x}_1^s\right), p\left(\boldsymbol{x}_0^s\right), p\left(\boldsymbol{x}_t^s|\boldsymbol{x}_0^s,\boldsymbol{x}_1^s\right)}\left\|v^{\mathrm{pos}}\left(\boldsymbol{x}_t^s, t, C^{\mathrm{rec}}| (1-b)\cdot c + b\cdot \emptyset \right)-\left(\boldsymbol{x}_1^s-\boldsymbol{x}_0^s\right)\right\|_2^2,
\end{equation}
where $b\sim \mathrm{Bernoulli}(p_{\mathrm{uncond}})$ indicates the probability to use the null condition. 

\section{Experimental Details}
\label{app: exp_details}
\subsection{Training and Sampling}
All experiments are implemented on 4 NVIDIA A100 GPUs. Specifically, we train SurfFlow for 320K iterations and set the initial learning rate of 5e-4. A plateau scheduler is used with a factor of 0.8 and patience of 10. The minimum learning rate is 5e-6. The batch size was 32 for each distributed node. An Adam optimizer is used with gradient clipping. A dropout ratio of 0.15 is adopted for the EGNN decoder. The weights for different loss components are set as $\lambda_{\mathrm{pos}}=0.2$, $\lambda_{\mathrm{ori}}=0.2$, $\lambda_{\mathrm{cat}}=1.0$, $\lambda_{\mathrm{con}}=1.0$, and $\lambda_{\mathrm{str}}=1.0$. 
For the encoder part, the residue embedding size is set as $\psi_{\mathrm{rec}} = \psi_{\mathrm{pair}} =128$. 
For the decoder part, the node and edge embedding sizes are set as 128 and 64 for IPA, respectively, while as 16 and 8 for EGNN, respectively, since the number of surface points $m + m'$ are orders of magnitude larger than the number of complex's residues $n_{\mathrm{pep}} + n_{\mathrm{rec}}$.  
We set the length of generated peptides the same as the length of their corresponding native peptides. We download the PepMerge data~\citep{li2024full} from its official repository:~\url{https://drive.google.com/drive/folders/1bHaKDF3uCDPtfsihjZs0zmjwF6UU1uVl}. 

\subsection{Physichemical Surface Features}
Three types of surface biochemical properties are leveraged in the experiments. To be specific, free electrons and potential hydrogen bond donors (FEPH) are categorical variables, while electrostatics and hydropathy are continuous variables. Therefore, $\psi_{\tau}=1$ and  $\psi_{\Upsilon}=2$. We resort to MaSIF~\citep{gainza2020deciphering}'s scripts to acquire these surface features.

\paragraph{Free electrons and proton donors.} The location of FEPH in the molecular surface was computed using a hydrogen bond potential as a reference. Vertices in the molecular surface whose closest atom is a polar hydrogen, a nitrogen, or an oxygen were considered potential donors or acceptors in hydrogen bonds. Then, a value from a Gaussian distribution was assigned to each vertex depending on the orientation between the heavy atoms. These initial values range from -1 (optimal position for a hydrogen bond acceptor) to +1 (optimal position for a hydrogen bond donor). Then the point is determined as an acceptor or a donor (a binary label) by whether FEPH is negative or positive. 

\paragraph{Hydropathy.} Each vertex was assigned a hydropathy scalar value according to the  Kyte and Doolittle scale of the amino acid identity of the atom closest to the vertex. These values, in the original scale, ranged between -4.5 (hydrophilic) to +4.5 (most hydrophobic) and were then normalized to be between -1 and 1. 

\paragraph{Poisson-Boltzmann continuum electrostatics.} PDB2PQR was used to prepare protein files for electrostatic calculations, and APBS (v.1.5) was used to compute Poisson-Boltzmann electrostatics for each protein. The corresponding charge at each vertex of the meshed surface was assigned using Multivalue, provided within the APBS suite. Charge values above +30 and below -30 were capped at those values, and then values were normalized between -1 and 1. 

\paragraph{Loss function for continuous surface features.} Compared to discrete physicochemical features, the loss is much easier to analyze forthe continuous type. Specifically, as the continuous feature values are normalized between -1 and 1, we leverage a normalized distribution $\mathcal{N}(0, 1)$ as the prior distribution. Correspondingly, with $\boldsymbol{\Upsilon}_i^s\sim \mathcal{N}(0, 1)$, the flow and vector field are:
\begin{align}
    \phi_t^{\mathrm{con}}\left(\boldsymbol{\Upsilon}_0^s, \boldsymbol{\Upsilon}_1^s\right)&=t \boldsymbol{\Upsilon}_1^s+(1-t) \boldsymbol{\Upsilon}_0^s, \\
    u_t^{\mathrm{con}}\left(\boldsymbol{\Upsilon}_t^s \mid \boldsymbol{\Upsilon}_1^s, \boldsymbol{\Upsilon}_0^s\right)&=\boldsymbol{\Upsilon}_1^s-\boldsymbol{\Upsilon}_0^s=\frac{\boldsymbol{\Upsilon}_1^s-\boldsymbol{\Upsilon}_t^s}{1-t}.
\end{align}
After that, a time-dependent translation-invariant surface network $v^{\text{con}}(\cdot)$ is used to predict the conditional vector field based on the current interpolant $\boldsymbol{\Upsilon}_t$ and the timestep $t$. The CFM objective of the surface continuous physicochemical features is formulated as:
\begin{equation}
\label{equ: cons_feat}
    \mathcal{L}_{\mathrm{con}}(\theta)=\mathbb{E}_{t\sim \mathcal{U}(0,1), p\left(\boldsymbol{\Upsilon}_1^s\right), p\left(\boldsymbol{\Upsilon}_0^s\right), p\left(\boldsymbol{\Upsilon}_t^s|\boldsymbol{\Upsilon}_0^s,\boldsymbol{\Upsilon}_1^s\right)}
    \left\|v^{\mathrm{con}}\left(\boldsymbol{\Upsilon}_t^s, t, C^{\mathrm{rec}}\right)-\left(\boldsymbol{\Upsilon}_1^s-\boldsymbol{\Upsilon}_0^s\right)\right\|_2,
\end{equation}

\subsection{Evaluation Metrics for Surface Consistency}
While structure consistency is considered in the metrics, it's unclear how the generated surface of generated protein-peptide structures is similar to the calculated surface of the generated protein-peptide structures. Inspired by recent molecular surface methods~\citep{sun2024dsr}, we utilize three evaluation metrics: IoU, CD, and NC, to measure the similarity between the molecular surfaces of designed and target proteins. For simplicity, all three metrics are normalized to a range of $0-1$. These metrics collectively offer a comprehensive assessment of the model's performance from multiple perspectives, as described below. 

\paragraph{IoU} The Intersection over Union (IoU) evaluates the overlap between the reconstructed volume and the ground truth shape (higher is better). For two arbitrary shapes $A, B \subseteq \mathbb{S} \in \mathbb{R}^{n}$, IoU is defined as $\text{IoU} = \frac{|A \cap B|}{|A \cup B|}$.

\paragraph{CD} The Chamfer Distance (CD) quantifies the discrepancy between two point sets $\mathcal{X}_{1}, \mathcal{X}_{2} \subset \mathbb{R}^{n}$ (lower is better). It is computed as:
$\mathrm{d}_{\mathrm{C}}\left(\mathcal{X}_{1}, \mathcal{X}_{2}\right) = \frac{1}{2} \left( \mathrm{d}_{\overrightarrow{\mathrm{C}}}\left(\mathcal{X}_{1}, \mathcal{X}_{2}\right) + \mathrm{d}_{\overrightarrow{\mathrm{C}}}\left(\mathcal{X}_{2}, \mathcal{X}_{1}\right) \right)$,
where $\mathrm{d}_{\overrightarrow{\mathrm{C}}}\left(\mathcal{X}_{1}, \mathcal{X}_{2}\right) = \frac{1}{|\mathcal{X}_{1}|} \sum_{\boldsymbol{x}_{1} \in \mathcal{X}_{1}} \min_{\boldsymbol{x}_{2} \in \mathcal{X}_{2}} \left\|\boldsymbol{x}_{1} - \boldsymbol{x}_{2}\right\|$.

\paragraph{NC} Normal Consistency (NC) evaluates the alignment of surface normals (higher is better). For two normalized unit vectors $\overrightarrow{n}_{i}$ and $\overrightarrow{m}_{j}$, the normal consistency is given by their dot product. To evaluate surface normals between the predicted surface points and normals $X_{\text{pred}} = \{(\boldsymbol{x}_{i}, \overrightarrow{n}_{i})\}$ and the ground truth surface points and normals $X_{\text{gt}} = \{(\boldsymbol{y}_{j}, \overrightarrow{m}_{j})\}$, the surface normal consistency $\Gamma$ is defined as:
$\Gamma\left(X_{\text{gt}}, X_{\text{pred}}\right) = \frac{1}{|X_{\text{gt}}|} \sum_{j \in |X_{\text{gt}}|} \left| \overrightarrow{n}_{j} \cdot \overrightarrow{m}_{\theta(\boldsymbol{y}_{j}, X_{\text{pred}})} \right|$, where $\theta(\boldsymbol{y}_{j}, X_{\text{pred}}) = \underset{i \in |X_{\text{pred}}|}{\arg \min} \left\|\boldsymbol{y}_{j} - \boldsymbol{x}_{i}\right\|_{2}^{2}$. 


\subsection{Flow Matching for Internal Structures} 
\label{app:fm_internal}
$\mathcal{L}_{\mathrm{str}}(\theta)$ in Equ.~\ref{equ:overall_loss} computes the loss for peptide's structures $p\left(\left\{\left(a_j, O_j, \boldsymbol{x}_j, \boldsymbol{\chi}_j\right)\right\}_{j=1}^{n_{\mathrm{pep}}} \big| C^{\mathrm{rec}}\right)$. Here, we provide a quick view of how to conduct full-atom CFM for residues' positions $\boldsymbol{x}$, types $a$, backbone torsions $O$, and side-chain angles $\boldsymbol{\chi}$. To be concrete, the objective is $\phi_t\left(\boldsymbol{x}_0, \boldsymbol{x}_1\right) = t\boldsymbol{x}_1 + (1-t)\boldsymbol{x}_0$ for residues' positions $\boldsymbol{x}$, $\exp_{O_0}\left(t \log _{O_0}\left(O_1\right)\right)$ for backbone orientations $O$, $\phi_t\left(\boldsymbol{\chi}_0, \boldsymbol{\chi}_1\right) = \left[t\boldsymbol{\chi}_1 + (1-t)\boldsymbol{\chi}_0\right]\mod 2\pi $ for side-chain angles $\boldsymbol{\chi}$, and $\phi_t\left(\Tilde{a}_0, \Tilde{a}_1\right) = t\Tilde{a}_1 + (1-t)\Tilde{a}_0$ for residue types $a$, where $\Tilde{a}$ is the representation of $a$ using a soft one-hot encodeing operation and satisfies $\mathrm{logit}(a_i) = \Tilde{a}_i\in \mathbb{R}^{20}$.

\section{Additional Results and Visualization}
\subsection{Surface Point Cloud Evolution}
\label{app:visual_evo}
We give some examples of how the surface point clouds move from time $t=0$ to the terminal time $t=1$. It can be seen that after time $t=0.5$, the point clouds start to form discernible shapes. 
\begin{figure}[ht]
    \centering
    \includegraphics[width=0.6\linewidth]{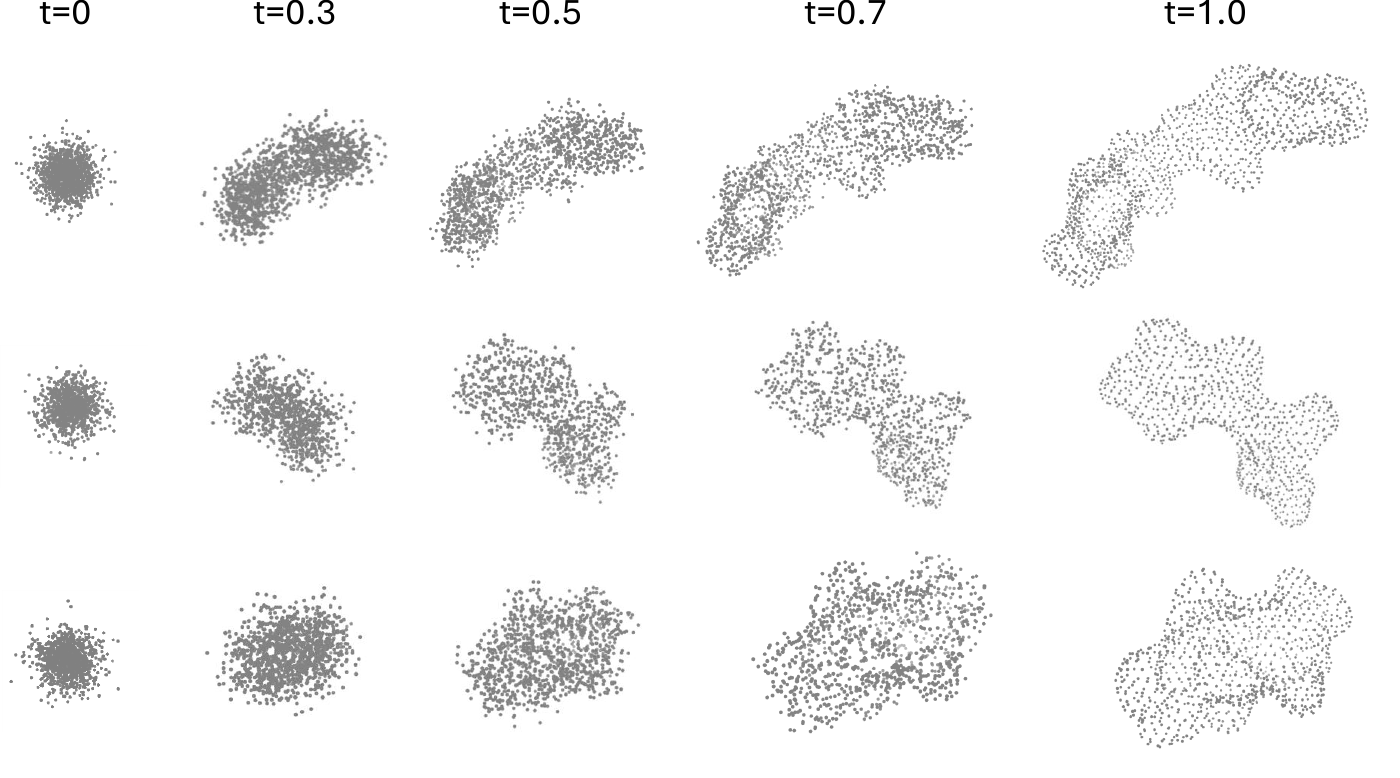}
    \caption{Visualization of the evolution of surface points' positions over time $t\in [0,1]$.}
    \Description{ }
    \label{fig:surface_noise}
\end{figure}

\subsection{Computational Efficiency Comparison}
As surface feature calculation may introduce computational overhead, we conducted a wall-time comparison of our approach with competing state-of-the-art methods, including RFDiffusion. The results, summarized in Tab.~\ref{tab:time}, reveal that while SurfFlow introduces a modest computational overhead due to the calculation of surface features, the additional time is well-justified by the improvements in design quality. 
Specifically, the surface feature extraction step accounts for approximately 12.8\% of the overall runtime, but this computation can be efficiently parallelized across multiple cores, minimizing its impact in practice. Moreover, it is worth noting that RFdiffusion was originally trained with 200 discrete timesteps. However, recent improvements enable reduced inference timesteps. RFdiffusion claims that running with as few as approximately 20 steps provides outputs of equivalent in silico quality to running with 200 steps (providing a 10X speedup) in many cases. But here, we still use 200 steps for a fair comparison. 
This analysis highlights the speed advantage of flow models, aligned with FrameFlow's discovery~\citep{yim2023fast}. It also indicates the trade-off between computational cost and the design quality of our SurfFlow, emphasizing the importance of feature-rich representations for advancing protein generative modeling. 
\begin{table}[ht]
    \caption{Comparison of inference time for different methods on the PepMerge test set. Each method is run in a single NVIDIA H100 GPU with 80GB of memory with the default setting. }
    \centering
    \resizebox{0.35\columnwidth}{!}{%
    \begin{tabular}{ccc} \toprule
     Model & Infer. Steps & Time (mins.)\\  \midrule
     RFdiffusion & 200 & 182.5\\
     PepFlow & 200 & 25.6 \\ 
     SurfFlow & 200 & 28.9 \\ \bottomrule
    \end{tabular}}
    \label{tab:time}
\end{table}

\section{Limitations and Future Works}
\label{app:feature_work}
Despite the enhancement of our SurfFlow over the original full-atom design mechanism, there is still room for future explorations. For instance, further improvements can be expected if the surface information of the receptor surface information is considered and incorporated into the joint distribution modeling. Namely, our objective becomes $C_{\mathrm{rec}}= \left\{a_j, O_j, \boldsymbol{x}_j, \boldsymbol{\chi}_j\right\}_{j=1}^{n_{\mathrm{rec}}} \cup \left\{\boldsymbol{x}_i^s, \boldsymbol{n}_i^s, \boldsymbol{\tau}_i^s, \boldsymbol{\Upsilon}_i^s\right\}_{i=1}^{m'}$, where $m'$ is the number of receptor's surface points. Moreover, the success of RFDiffusion implies that pretraining on regular proteins in PDB can be beneficial.

\end{document}